\documentclass{fairmeta}

\usepackage[utf8]{inputenc}
\usepackage{url}
\usepackage{amsfonts}
\usepackage{amsmath}
\usepackage{amssymb}
\usepackage{nicefrac}
\usepackage{wrapfig}
\usepackage{tikz}
\usepackage{float}
\usetikzlibrary{arrows.meta, positioning, fit, backgrounds, shapes.geometric, calc}

\newcommand{\ie}{i.e., }
\newcommand{\eg}{e.g., }

\title{Boosting Brain-to-Image Decoding with TRIBE~v2 Data Augmentation}
\author{Yohann Benchetrit}
\author{Marlène Careil}
\author{Simon Dahan}
\author{Hubert Banville}
\author{Stéphane d'Ascoli}
\author{Jean-Rémi King}
\affiliation{Meta AI}
\correspondence{Yohann Benchetrit at \email{ybenchetrit@meta.com}}

\abstract{%
Brain decoding is limited by the availability of labeled neural data, and remains challenging in low-data regimes. To address this issue, we investigate whether and when brain decoding can be boosted by augmenting small fMRI datasets with synthetic data generated by a pretrained model of fMRI responses to stimuli. We use TRIBE~v2, a large encoding model pretrained on more than 1000\,hours of fMRI responses to video, audio and language. For each dataset, we evaluate systematic grids that show how the performance of image decoders varies with the amount of synthetic data used for training. Our results, based on two datasets (the 7T fMRI Natural Scenes Dataset and 3T fMRI BOLD5000), show up to 68\% improvement in Top-10 image-retrieval accuracy compared to decoders trained only on real data. Importantly, the proportion of augmented data required to reach a given image decoding performance needs to be adjusted depending on the data source. 
Surprisingly, image decoders trained exclusively on synthetic fMRI can perform above chance in some settings, suggesting that TRIBE~v2 can support zero-shot brain-to-image decoding.
Together, these results show how large-scale models of the fMRI responses to sight, sound and language may provide a foundation to improve the data efficiency for image decoding.}

\begin{document}
\maketitle

\section{Introduction}
\label{sec:intro}

Brain-to-image decoding has progressed rapidly thanks to the pairing of fMRI with powerful pretrained visual representations and generative models~\citep{scotti2024mindeye2,ozcelik2023brain,chen2023seeing,benchetrit2024brain,shen2019end}.
Yet its practical reach remains restricted by the amount (and quality) of brain imaging data:
High-performing decoders typically require thousands of stimulus-response pairs from the same individual, collected over many scanning sessions~\citep{banville2025scaling}.
For example, one of the most popular datasets for fMRI-to-image reconstruction is 
the Natural Scenes Dataset (NSD), which contains 30-40 hours of 7T fMRI per subject~\citep{allen2022massive}. In comparison, most neuroscience studies are far smaller, noisier, and heterogeneous e.g.~\citep{chang2019bold5000,shen2019end,hebart2023things}.
This discrepancy creates a main bottleneck: modern decoding methods are strongest precisely in the data regime that few laboratories can afford to collect.

Data augmentation is the standard response to data scarcity in machine learning, but fMRI presents an unusual challenge.
Classical augmentations such as cropping, flipping, or color jitter do not apply trivially to neural recordings, because neural data is inherently tied to the fixed biological architecture of the subject rather than the fluid, translation-invariant structure of a 2D image grid. In addition, `model-free' alternatives like noise injection  or trial-mixing strategies like MixCo ~\citep{scotti2024mindeye,scotti2024mindeye2}  can regularize a decoder and improve performance, but they operate on neural responses that have already been measured and therefore do not provide responses for new external stimuli. 

Here, we investigate an alternative -- model-based \textit{and} stimulus-conditioned -- data augmentation: we leverage a pretrained encoding
model to predict how the brain would respond to new stimuli, then use those predicted responses as additional training data for the inverse problem \ie decoding.

We benchmark this approach on fMRI-to-image decoding using TRIBE~v2~\citep{dascoli2026tribe}, a state-of-the-art foundation model trained to predict fMRI responses to video, audio, and language. Although TRIBE~v2 was not trained specifically on static image stimuli, its visual pathway can be applied to images by treating them as short static videos. We use TRIBE~v2 to synthesize fMRI responses for novel images, train fMRI-to-image decoders on controlled mixtures of real (\ie measured) and synthetic fMRI, and evaluate retrieval performance on a held-out set of real fMRI data.

Beyond investigating \emph{whether} synthetic fMRI helps image decoding, we here aim to resolve a more practical question: \emph{under what conditions and data regimes can an fMRI-to-image decoder leverage synthetic data?}

By varying two quantities -- the fraction of real fMRI and the amount of synthetic, TRIBE-generated fMRI added relative to that real subset -- we show that using synthetic fMRI data can substantially improve image decoding in low-to-medium data regimes, but excessive synthetic data can saturate or even hurt performance. We find that the best real-to-synthetic ratio differs across datasets and decoders.

This leads to three main contributions:
\begin{enumerate}
    \item We propose and characterize a subject- and dataset-agnostic, image-conditioned data augmentation pipeline for fMRI-to-image decoding. We use TRIBE~v2's visual pathway to synthesize fMRI responses to novel images, without fitting or adapting the model to the decoded dataset or subject.
    \item We apply this protocol to the Natural Scenes Dataset and BOLD5000, and grid how linear and deep decoders respond to different real-to-synthetic training ratios. In low-data regimes, adding TRIBE-synthetic responses improves Top-10 image-retrieval accuracy by up to 68\%. We further show that the same augmentation strategy can improve an fMRI-to-image reconstruction model~\citep{careil2025dynadiff}.
    \item We show that TRIBE augmentation is an empirical tool that requires calibration rather than a plug-and-play method. Its effect depends on the dataset, decoder capacity, and augmentation factor, yielding gains in selected regimes but saturation or decline at higher factors.
\end{enumerate}

\section{Related Work}
\label{sec:related}

\paragraph{Brain decoding.}
Brain decoding aims to infer perceived or imagined content from neural activity.
Early fMRI work decoded simple visual features and natural image identity with linear and Bayesian models~\citep{kamitani2005decoding,kay2008identifying,naselaris2009bayesian}.
Recent systems use pretrained visual representations and diffusion models to retrieve or reconstruct images from brain activity, 
including MindEye~\citep{scotti2024mindeye}, MindEye2~\citep{scotti2024mindeye2}, Brain-Diffuser~\citep{ozcelik2023brain}, MindVis~\citep{chen2023seeing}, and DynaDiff~\citep{careil2025dynadiff}.
As shown in this prior work, these advances make the data bottleneck more visible: strong decoders usually require many subject-specific samples.
Where MindEye2 addresses this challenge through multi-subject pretraining, we explore a complementary approach: using image-conditioned fMRI synthesis to expand the decoder's training set without acquiring additional real fMRI.

\paragraph{fMRI encoding models.}
Encoding models predict neural responses from stimuli and have long served as a bridge between computational neuroscience and machine learning~\citep{kay2008identifying,naselaris2011encoding,st2023brain}.
Modern encoding models use deep visual features and can predict responses across the visual cortex~\citep{st2023brain,conwell2023can}.
TRIBE~v2 extends this line of research to a large-scale, tri-modal model trained on over 1{,}000 hours of fMRI from 720 subjects~\citep{dascoli2026tribe}.
Importantly, we use TRIBE~v2 not as a forward model, but as a generator of synthetic fMRI responses for the inverse task, \ie decoding.

\paragraph{Synthetic data for neuroimaging and machine learning.}
Synthetic data augmentation improves data efficiency in image classification, medical imaging, and language modeling~\citep{azizi2023synthetic,he2022synthetic,fernandez2022can,scotti2024mindeye,scotti2024mindeye2}.
In fMRI, augmentation typically relies on noise injection, trial manipulation, or generative models trained on the target data~\citep{nguyen2023blends,wang2023learning}.
Recent work has also explored image-conditioned fMRI synthesis.
MindSimulator~\citep{bao2025mindsimulator} learns subject-specific generative encoding models from measured NSD data and applies the generated responses to concept-selective cortical localization, with validation against functional localizers.
Here, we study synthetic fMRI in a different setting.
An externally pretrained population-level model generates additional training data for subject-specific decoders, without fitting or adaptation to the target dataset or subject.
We evaluate these decoders on held-out measured fMRI and map how performance varies across real-to-synthetic ratios, datasets, and decoder families.

\section{Methods}
\label{sec:method}

\begin{figure}[t]
  \centering

  \begin{subfigure}[t]{0.98\linewidth}
    \centering
    \resizebox{\linewidth}{!}{%
    \begin{tikzpicture}[
        font=\small,
        node distance=0.85cm and 1.05cm,
        box/.style={draw, rounded corners=3pt, very thick, align=center,
          minimum height=1.0cm, minimum width=2.9cm, inner sep=4pt},
        data/.style={box, fill=blue!7, draw=blue!45!black},
        synth/.style={box, fill=teal!8, draw=teal!55!black},
        model/.style={box, fill=orange!9, draw=orange!60!black},
        eval/.style={box, fill=purple!7, draw=purple!55!black},
        arrow/.style={-{Latex[length=2mm]}, very thick},
        smallbox/.style={draw, rounded corners=2pt, align=center,
          minimum height=0.62cm, inner sep=3pt}
      ]

      \node[synth] (pool) {Unseen COCO images};
      \node[model, right=of pool] (tribe) {TRIBE~v2\\image $\rightarrow$ fMRI};
      \node[synth, right=of tribe] (synthetic) {Synthetic training pairs\\$a \times pN$};

      \node[model, below=of synthetic] (train)
        {Train decoder\\fMRI $\rightarrow$ DINOv2-small\\$(1+a)pN$ training pairs};
      \node[data] (real) at (pool |- train) {Real training pairs\\$pN$};
      \node[eval, right=of train] (test)
        {Held-out real fMRI\\Top-10 image retrieval};

      \draw[arrow] (pool) -- (tribe);
      \draw[arrow] (tribe) -- (synthetic);
      \draw[arrow] (real) -- (train);
      \draw[arrow] (synthetic) -- (train);
      \draw[arrow] (train) -- (test);

    \end{tikzpicture}%
    }
    \caption{TRIBE~v2 image-conditioned fMRI augmentation protocol.}
  \end{subfigure}

  \vspace{0.35em}

  \begin{subfigure}[t]{0.82\linewidth}
    \centering
    \includegraphics[width=\linewidth]{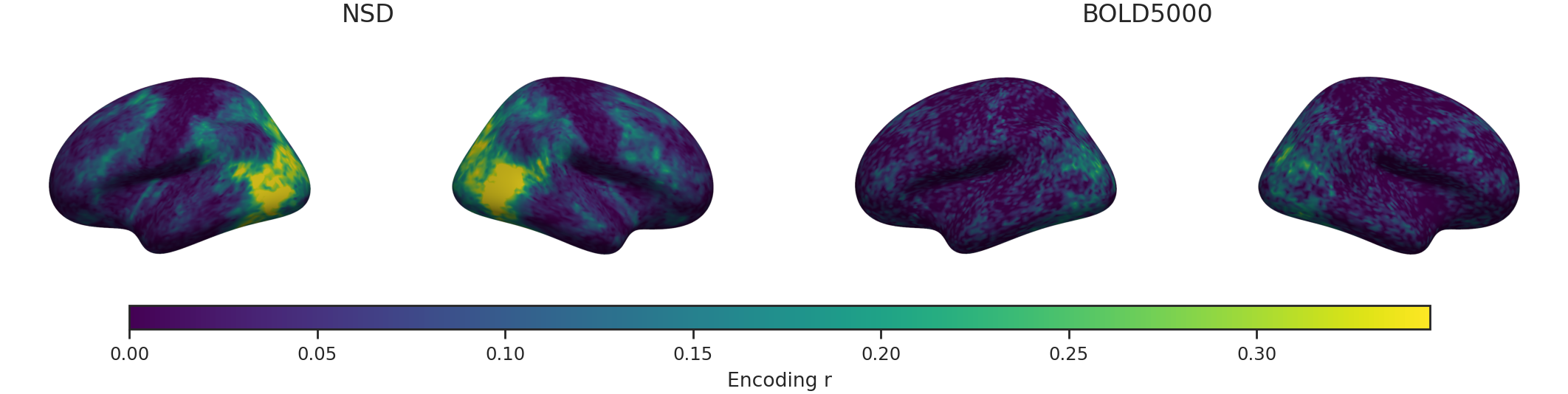}
    \caption{DINOv2-small image embeddings predict cortical fMRI responses.}
  \end{subfigure}

  \caption{\textbf{TRIBE~v2 image-conditioned fMRI augmentation and target-embedding neural signal.}
  \textbf{(a)} For each subject and dataset, we retain $pN$ real image-fMRI training pairs and sample additional COCO images not seen by that subject.
  TRIBE~v2 predicts synthetic fMRI responses to these images, yielding $apN$ synthetic training pairs, where $a$ is the augmentation factor.
  We train an fMRI-to-DINOv2-small decoder on the resulting mixture of real and synthetic pairs and evaluate it only on held-out real fMRI using Top-10 image retrieval.
  \textbf{(b)} Brain-encoding grids show Pearson correlation between DINOv2-small-predicted and real fMRI cortical activations on fsaverage5, averaged across subjects within each dataset, supporting the use of DINOv2-small as the decoded image representation.}
  \label{fig:method}
\end{figure}

\subsection{TRIBE~v2 as a Synthetic fMRI Generator}
\label{sec:tribe}

TRIBE~v2~\citep{dascoli2026tribe} predicts BOLD fMRI responses on the cortical surface (fsaverage5~\citep{fischl1999high}) from video, audio, and language inputs.
It uses frozen modality-specific backbones, including V-JEPA~2~\citep{assran2025vjepa2} for visual input (videos), and integrates their features with a transformer encoder to predict responses of $T=100~s$ at a frequency of 1~Hz (\ie TR=1~s), starting 5~s after stimulus onset (to account for hemodynamic delay). We use TRIBE~v2 in its default mode, which predicts population-level responses without adapting to a target subject. For brevity, we refer to this model as ``TRIBE'' in the remainder of the paper.

\paragraph{Synthesizing fMRI for images.} TRIBE was trained on naturalistic video, audio, and text, and not on isolated still images.
To apply it to static image datasets, we therefore convert each image into a short 3~s still video, where every frame shows the same image. Of note, this use of TRIBE is significantly \textit{out-of-distribution}, as we feed it video that has no natural motion; a property absent from TRIBE's training set.
The resulting synthetic fMRI time-series of $T$~seconds is resampled to match the target dataset acquisition frequency and truncated to a single timestep (\ie 1~TR). 

\subsection{Data-Augmentation Operating Grids for Image Decoding}
\label{sec:framework}

\paragraph{Image Decoding.} Let $\mathcal{D}=\mathcal{D}_{\mathrm{train}}\cup\mathcal{D}_{\mathrm{test}}$ be a dataset of real fMRI responses to images for a given subject, with disjoint train and test splits. That is, $\mathcal{D}_{\mathrm{train}}=\{(s_i,y_i)\}_{i=1}^{N}$ and $\mathcal{D}_{\mathrm{test}}=\{(s_i,y_i)\}_{i=N+1}^{N+N_{\mathrm{test}}}$, where $s_i$ is an image stimulus and $y_i \in \mathbb{R}^{V}$ is the corresponding fMRI brain response from the subject. The task of image decoding is to train a model to predict a representation (or reconstruction) of $s$ from $y$.

\paragraph{Data Augmentation.} Let $f_\theta(s)$ denote the TRIBE-synthesized fMRI response to stimulus $s$.
For a \textit{percentage of retained real-data} $p$ and \textit{TRIBE augmentation factor} $a \in \mathbb{R}_{\geq 0}$, we train a model $\mathcal{M}$ on
\begin{equation}
    \mathcal{D}_{p,a}
    = \{(s_i,y_i): i \in \mathcal{I}_p\}
    \cup
    \{(\tilde{s}_j, f_\theta(\tilde{s}_j)): j=1,\ldots, \lfloor a|\mathcal{I}_p| \rfloor\},
\end{equation}
where $\mathcal{I}_p$ is a random subset of $\mathcal{D}_{\mathrm{train}}$ of size $\lfloor pN \rfloor$, and the $\tilde{s}_j$ are sampled from an image augmentation pool $\mathcal{A}$ disjoint from $\mathcal{D}$. We evaluate $\mathcal{M}$ on $\mathcal{D}_{\mathrm{test}}$, which is kept fixed and never augmented.
The condition $a=0$ means subsampling a subset of $\mathcal{D}_{\mathrm{train}}$ of $p$\% of its total size, and represents a setting where the availability of real data is limited, with no access to synthetic fMRI augmentation. We refer to this as the \textit{matched real-only} setting (at $p$\%).

\paragraph{Synthetic-only Image Decoding.} On the other hand, we train 'real-data zero-shot', synthetic-only image decoders: the model is trained only on TRIBE-synthesized responses and evaluated on the real fMRI in $\mathcal{D}_{\mathrm{test}}$.

\paragraph{Operating Grids.} By training decoders over representative values of $p$ and $a$ and evaluating them on the same held-out dataset $\mathcal{D}_{\mathrm{test}}$ of real data, we obtain an \textit{operating grid} of how the decoding performance $\gamma(p,a)$ of models varies (on $\mathcal{D}_{\mathrm{test}}$) with the amount of real data available ($p$) and the amount of synthetic TRIBE data added ($a$).

\paragraph{Encoder-decoder Leakage.} 
We use the output of the last layer of DINOv2-small~\citep{oquab2023dinov2} as the decoding target. While it does not entirely rule out all forms of representational overlap with TRIBE's visual encoder (V-JEPA~2), it intentionally reduces a direct leakage concern: that augmentation gains could arise simply from using the same visual features to synthesize fMRI and to score the decoder.

\subsection{Reconstructing Images from BOLD fMRI}
\label{sec:dynadiff-method}
Beyond the task of decoding an image embedding, we use DynaDiff~\citep{careil2025dynadiff} to test whether synthesizing fMRI responses to images with TRIBE also helps reconstruct images from fMRI responses. 

We train DynaDiff on $\mathcal{D}_{p,a}$ (Section~\ref{sec:framework}) for $p=100$\% and $a > 0$. We add a modality-specific linear layer at the brain module's input, which tells the model whether it is processing real or synthetic data and facilitates convergence. Apart from this layer and minor adaptations of the temporal aggregation, the model is shared across modalities. Training otherwise follows \citet{careil2025dynadiff}; details are in Appendix~\ref{app:dynadiff}.

\subsection{Models and Evaluation}
\label{sec:decoding}

\paragraph{Linear decoders.} Our primary decoder is a ridge regression model that maps fMRI inputs (\ie cortical brain activations) to image embeddings.
This pipeline normalizes fMRI features into standard units (across training samples and voxels) and fits a ridge regression with per-target regularization and Leave-One-Out Cross-Validation.

\paragraph{Deep decoders.} We additionally evaluate a MindEye-style residual MLP decoder~\citep{scotti2024mindeye}.
This model maps fMRI cortical responses to image embeddings through a residual MLP; architecture and optimization details are provided in Appendix~\ref{app:fmrimlp-architecture}.
We train it with a CLIP contrastive retrieval loss as in \citep{benchetrit2024brain,banville2025scaling}: for each batch, the predicted image embedding is matched to its paired target against the other batch targets using cross-entropy over dot-product similarities.
Target embeddings are normalized in the loss, with fixed temperature $\tau=1$. The CLIP loss is one-way \ie it is not symmetrized across the target-to-prediction direction.
Both decoders are trained separately for each target subject.

\paragraph{Decoding Metrics.} The primary metric is Top-$K$ image retrieval accuracy on the entire test set (made only of real data).
Given a predicted image embedding, we rank all candidate test images by cosine similarity to their ground-truth embeddings and score whether the correct image appears among the top $K$.
We also report median retrieval rank in the Appendix, defined as the median of rank positions of the correct image in the same similarity-sorted retrieval list; lower ranks indicate better performance. 

\paragraph{Reconstruction Metrics.} Following standard practice in fMRI-to-image reconstruction~\citep{scotti2024mindeye2}, we report PixCorr and SSIM for low-level similarity; AlexNet-2, AlexNet-5, InceptionV3, and CLIP for feature-space similarity; and EfficientNet and SwAV distances for high-level semantic similarity.

\section{Experiments and Results}
\label{sec:results}

\subsection{Datasets}
\label{sec:datasets}

\paragraph{Hemodynamic delay.} We evaluate our approach on two image-fMRI datasets.
For both datasets, we decode a one-TR response window at the expected BOLD peak, starting 5~s after stimulus onset (consistent with TRIBE predictions), and train separate decoders for each subject.

\paragraph{Natural Scenes Dataset.}
NSD~\citep{allen2022massive} is a 7T dataset acquired with whole-brain gradient-echo EPI at 1.8 mm isotropic resolution and TR=1.6 s.
It contains eight participants, each scanned for 30--40 one-hour sessions while viewing natural images from MS-COCO.
Following common practice for image-decoding work, we use the four subjects who completed the full 40-session protocol and have fsaverage-space responses in our pipeline (subjects 1, 2, 5, and 7).
Each of these subjects viewed 10{,}000 unique images, repeated three times; the standard split uses 9{,}000 subject-specific training images and a shared 1{,}000-image test set.
Images were presented for 3 s with a 1-s blank interval.

\paragraph{BOLD5000.}
BOLD5000~\citep{chang2019bold5000} is a smaller 3T image-fMRI dataset acquired with multiband T2*-weighted EPI at 2 mm in-plane resolution and TR=2 s.
It includes four participants, each scanned across repeated 1.5-hour sessions, with one participant completing fewer sessions than the others.
Stimuli were drawn from MS-COCO, ImageNet, and SUN scene images, giving the full-session subjects 4{,}916 unique images.
A small repeated subset of 112 images forms the held-out test set, and each stimulus was shown for 1 s followed by a 9-s fixation interval.

\paragraph{Augmentation pool.} For each subject, the synthetic augmentation pool is sampled without replacement from COCO~\citep{lin2014microsoftcococommonobjects}, after excluding by COCO ID all images seen by that subject in either the training or test set.
For NSD, this excludes at least the 10{,}000 images seen by each subject.
For BOLD5000, we exclude all 2{,}000 of the dataset's stimuli drawn from COCO.
As such, the augmentation pool has no exact image overlap with the training or test sets of our decoders.
Moreover, neither NSD nor BOLD5000 are part of TRIBE's training data~\citep{dascoli2026tribe}.

\paragraph{Single-trial responses.} Some stimuli are presented multiple times to subjects in the training set, most notably in NSD where each image is shown three times.
Before constructing the operating-grid conditions, we keep one randomly selected fMRI response per subject and image and discard all other repetitions, and apply this deduplication uniformly to all real-only, TRIBE, and control scenarios in both datasets.
This avoids an imbalance between real data, which can contain multiple noisy repetitions of the same stimulus, and TRIBE predictions, which are deterministic and therefore provide only one synthetic response per image.

\paragraph{fMRI preprocessing.} Anatomical and functional MRI data are preprocessed with fMRIPrep~\citep{esteban2019fmriprep} using its default workflow, requesting cortical-surface outputs in fsaverage space.
We retain the fsaverage5 time series from each run so that real fMRI and TRIBE predictions live on the same 20{,}484-vertex cortical mesh.

\subsection{Experimental Setup}
\label{sec:setup}

\paragraph{Operating Grids.} We compute operating grids $\gamma(p,a)$ for each dataset and each decoder family, using real-data percentages $p \in \{10,30,50,70,90,100\}\%$ and augmentation factors $a \in \{0,0.25,0.5,0.75,1,2,4,8,16\}$.
For each dataset $\mathcal{D}$, a decoder is trained separately for each subject on $\mathcal{D}_{p,a}$ and is evaluated on the fixed held-out test set $\mathcal{D}_{\mathrm{test}}$ of this subject (\ie our decoders are \textit{single-subject}). We pick $\mathcal{D}_{\mathrm{test}}$ as defined by the dataset itself, following previous Image Decoding studies (see Section~\ref{sec:datasets}).
The resulting grids summarize performance both as a percentage of the 100\% real-data reference (Panel A in Figures~\ref{fig:ridge-operating} and \ref{fig:fmrimlp-operating}) and as the relative gain over the matched real-only baseline ($a=0$)  at the same retained-real percentage $p$ (Panel B in Figures~\ref{fig:ridge-operating} and \ref{fig:fmrimlp-operating}). Scores are computed after averaging 5 seeds within subject, controlling dataset subsampling and train / validation splits. 

\paragraph{Models.} We train our decoders to predict a DINOv2-small embedding of a stimulus image from an input of one TR of cortical fsaverage5 activations at stimulus onset 5~s (totalizing 20{,}484-vertex values across both hemispheres). Ridge regressions are tuned over 20 alphas evenly log-spaced from $1$ to $10^8$. The deep models use a residual architecture as introduced by \citet{scotti2024mindeye} (see Appendix~\ref{app:fmrimlp-architecture} for details) and are trained using AdamW with learning rate $5 \times 10^{-4}$ and weight decay $0.01$. Optimization uses a step-wise OneCycleLR schedule with maximum learning rate $10^{-3}$ and warm-up over the first $10\%$ of training steps. Models are trained for up to 40 epochs with early stopping on validation loss (20\% validation split) and patience 10.

\paragraph{Recording times.} The percentages can also be read as approximate scan-time budgets (Appendix~\ref{app:scan-time}).
For NSD, the 100\% real reference corresponds to 9{,}000 training images, or roughly 10 hours of image-task acquisition per subject (3 s presentation + 1 s blank interval).
Thus $p=\{10,30,50,70,90,100\}\%$ corresponds to about $\{1,3,5,7,9,10\}$ real hours.
For BOLD5000, the approximately 4{,}804 non-test images are presented for 1 s followed by 9 s fixation, so the full real training set corresponds to roughly 13.3 hours per subject; the same $p$ values correspond to about $\{1.3,4.0,6.7,9.3,12.0,13.3\}$ hours.

\subsection{Operating Grids for Ridge decoders}
\label{sec:ridge-results}

\begin{figure}[t]
  \centering
  \begin{subfigure}[t]{0.49\linewidth}
    \centering
    \includegraphics[width=\linewidth]{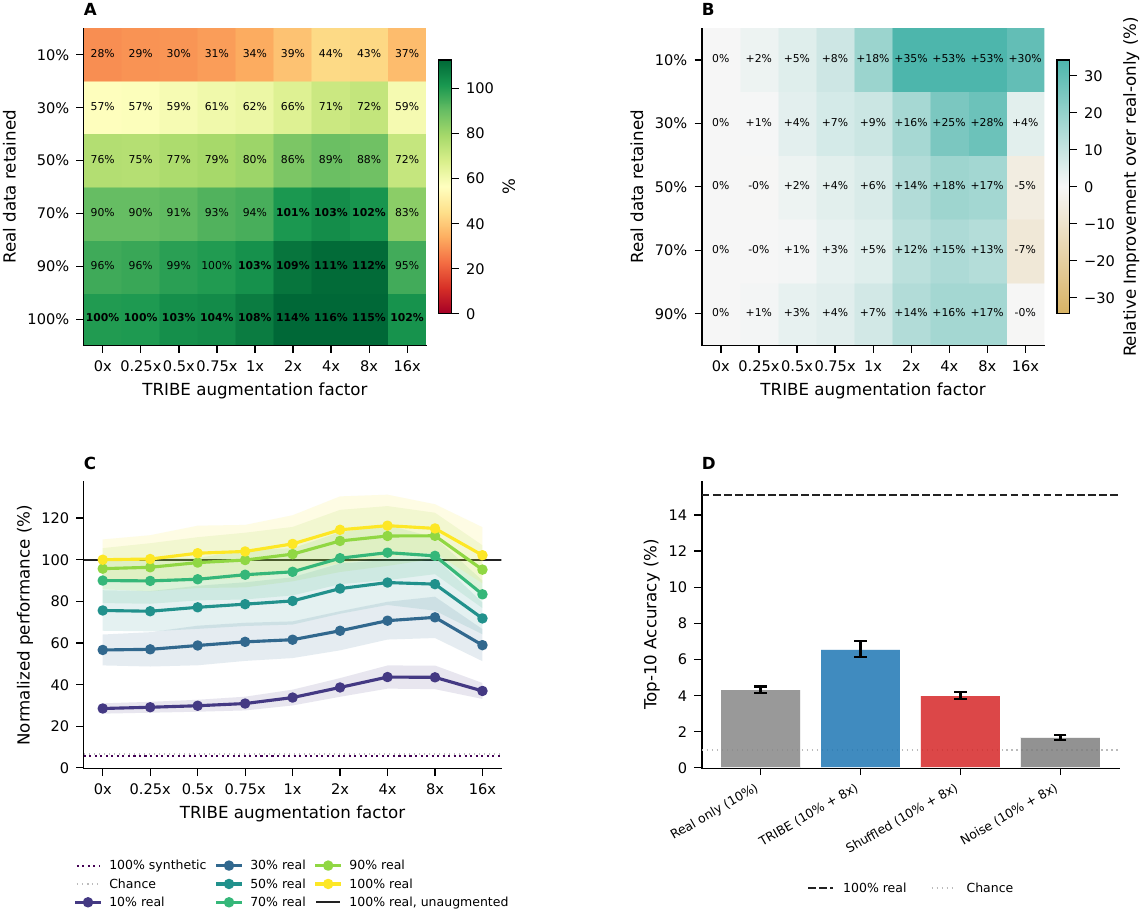}
    \caption{NSD}
  \end{subfigure}
  \hfill
  \begin{subfigure}[t]{0.49\linewidth}
    \centering
    \includegraphics[width=\linewidth]{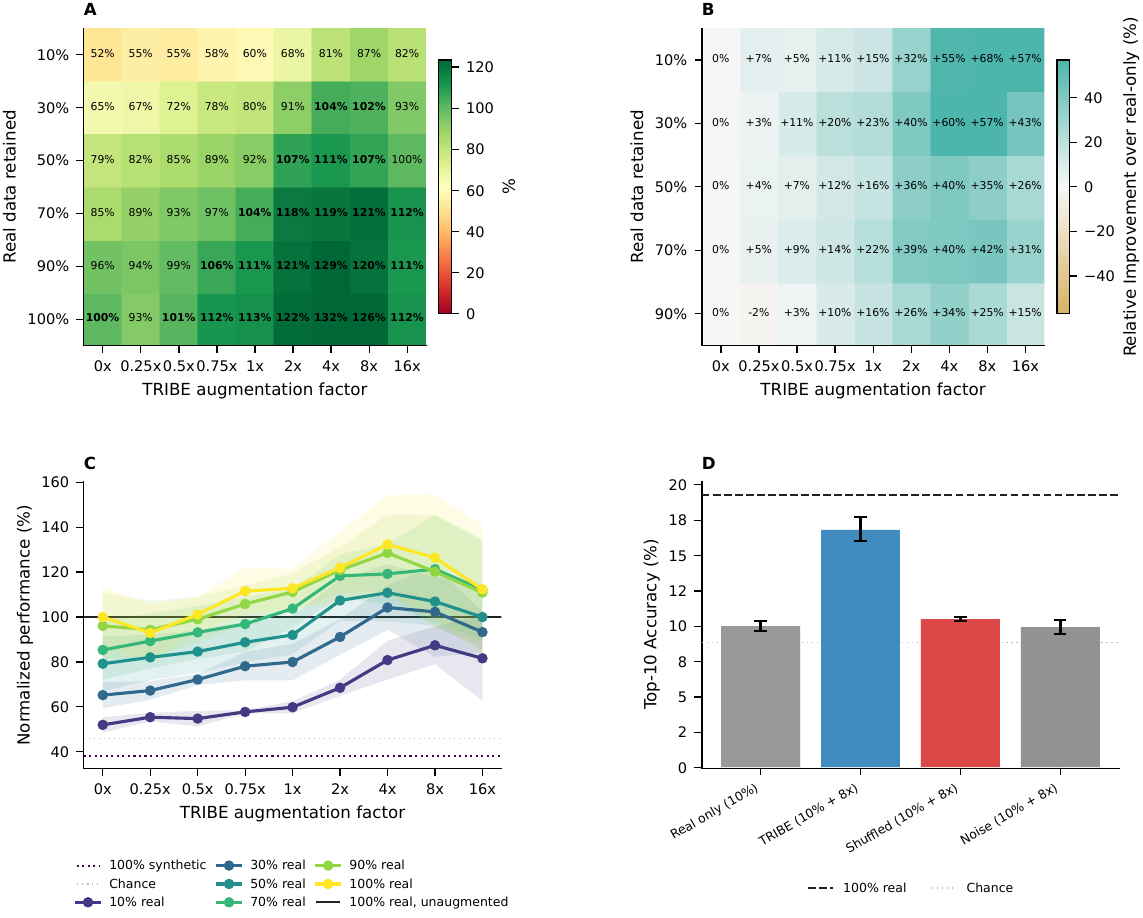}
    \caption{BOLD5000}
  \end{subfigure}
  \caption{\textbf{Operating grids for TRIBE augmentation using Ridge decoders.}
  \textbf{A}. Top-10 retrieval accuracy normalized to the performance of a 100\% real-data-only baseline (\ie a ridge model trained on the full, unaugmented training set); values above 100\% indicate that TRIBE augmentation surpasses this 100\% real-data-only baseline.
  \textbf{B}. Relative performance improvement over the matched real-only   condition ($a=0$) at the same retained-real percentage $p$\%.
  \textbf{C}. Augmentation scaling curves for each $p$\% of real-data retained, including chance and synthetic-only baselines. Shading shows SEM across all subjects (four subjects for each dataset).
  \textbf{D}. Raw Top-10 retrieval at the selected real-data fraction $p$\%, comparing TRIBE with matched real-only, Shuffled, and noise-augmentation controls; the dashed line marks the full real-only reference ($p=100\%$).
  Ridge models provide the cleanest operating-grid evidence: TRIBE augmentation improves low- and medium-data regimes. The best factor depends on dataset and the fraction of real data.}
  \label{fig:ridge-operating}
\end{figure}

Figure~\ref{fig:ridge-operating} shows operating grids, augmentation scaling curves and controls averaged across all subjects of each dataset (shading in Panel C and error bars in Panel D indicate SEM across subjects).
On the Natural Scenes Dataset (NSD), augmenting data with TRIBE produces substantial gains in low-data regimes. For example: around 90\% of the full real-only dataset performance (\ie $p=100$\%, $a=0$) can be obtained with only 50\% of retained real data (see $p=50$\%, $a=4$). 
In scan-time terms, because the deduplicated NSD training set corresponds to about 10 hours of real fMRI per subject, this means reaching roughly 90\% of full-data performance with about 5 real hours instead of 10. The full-real-only baseline's performance can be reached with $70$\% of real data, or about 7 real hours, saving roughly 3 hours per subject. Furthermore, TRIBE provides relative improvements up to 53\% even in the lowest regime $p=10$\%. 

At larger retained-real fractions, the relative improvements become smaller and can saturate, indicating that TRIBE augmentation is most useful when real data is scarce. Similarly, adding too much new data eventually saturates or hurts performance (topping around $a=8$). Of note, the 100\% synthetic-only data reference is at chance level and shows that real data remains necessary to decode NSD.

The BOLD5000 dataset shows even stronger performance gains. The Ridge decoders benefit substantially from TRIBE at $p=10\dots50$\% of retained real data, reaching already approximately 90\% of the full real-only baseline performance for $p=10$\% and 100\% with only $p=30$\% of real data retained. In scan-time terms, this corresponds to about 4.0 hours of real BOLD5000 training data instead of 13.3 hours, a saving of roughly 9 hours per subject. The gain brought by adding TRIBE synthetic data remains consistent for all presented data regimes and saturates between augmentation factors $a=4$ and $a=8$, at a slightly smaller augmentation factor than NSD. Also, in contrast to NSD, where the 100\%-synthetic-only baseline is roughly at chance, the corresponding BOLD5000 reference is slightly above, suggesting that TRIBE's subject-agnostic synthetic responses carry some transferable visual signal in this specific dataset.

Remarkably, despite the different acquisition protocols and absolute performance levels, the strongest relative-improvement cell for Ridge appears at the same operating point in both datasets: $p=10\%$ retained real data and augmentation factor $a=8$. See Appendix~\ref{app:additional-metrics} for per-subject Top-10 operating grids and subject-averaged median-rank grids for Ridge decoders.

\begin{figure}[t]
  \centering
  \begin{subfigure}[t]{0.49\linewidth}
    \centering
    \includegraphics[width=\linewidth]{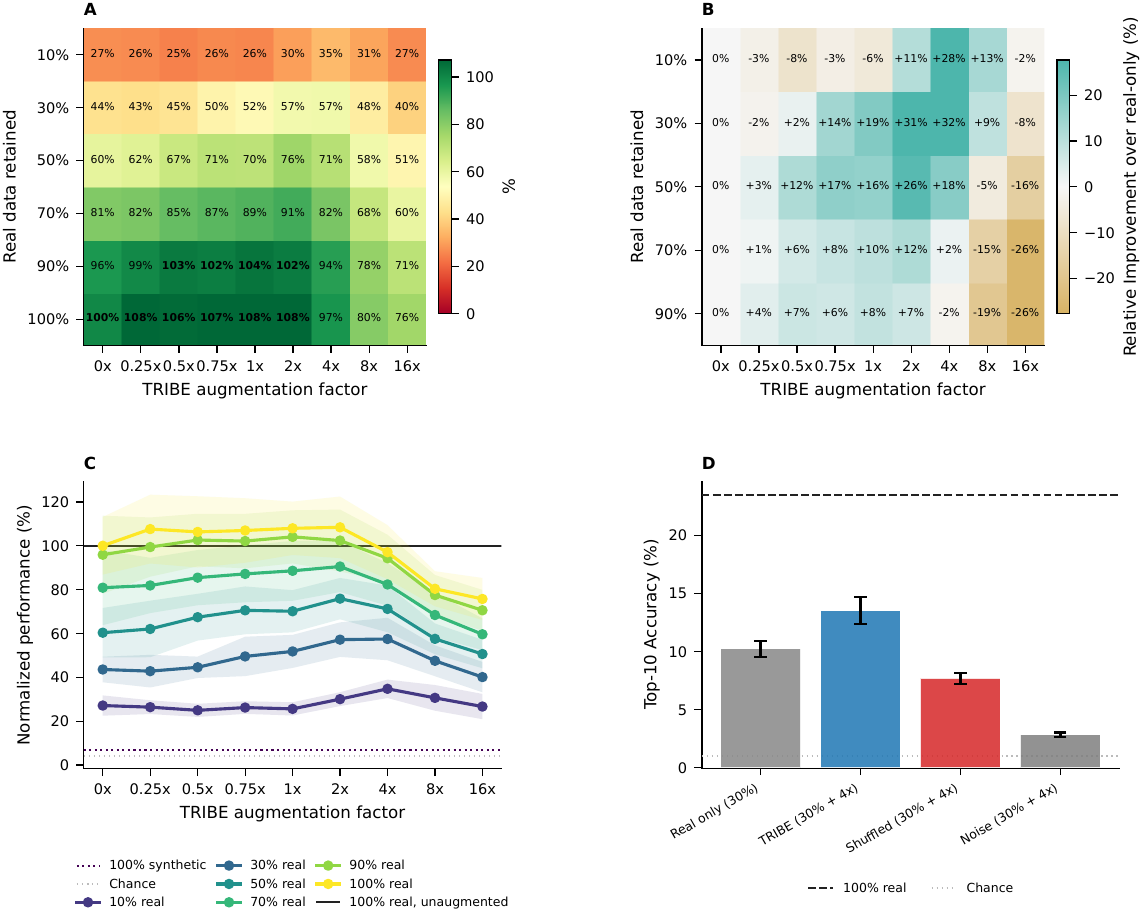}
    \caption{NSD}
  \end{subfigure}
  \hfill
  \begin{subfigure}[t]{0.49\linewidth}
    \centering
    \includegraphics[width=\linewidth]{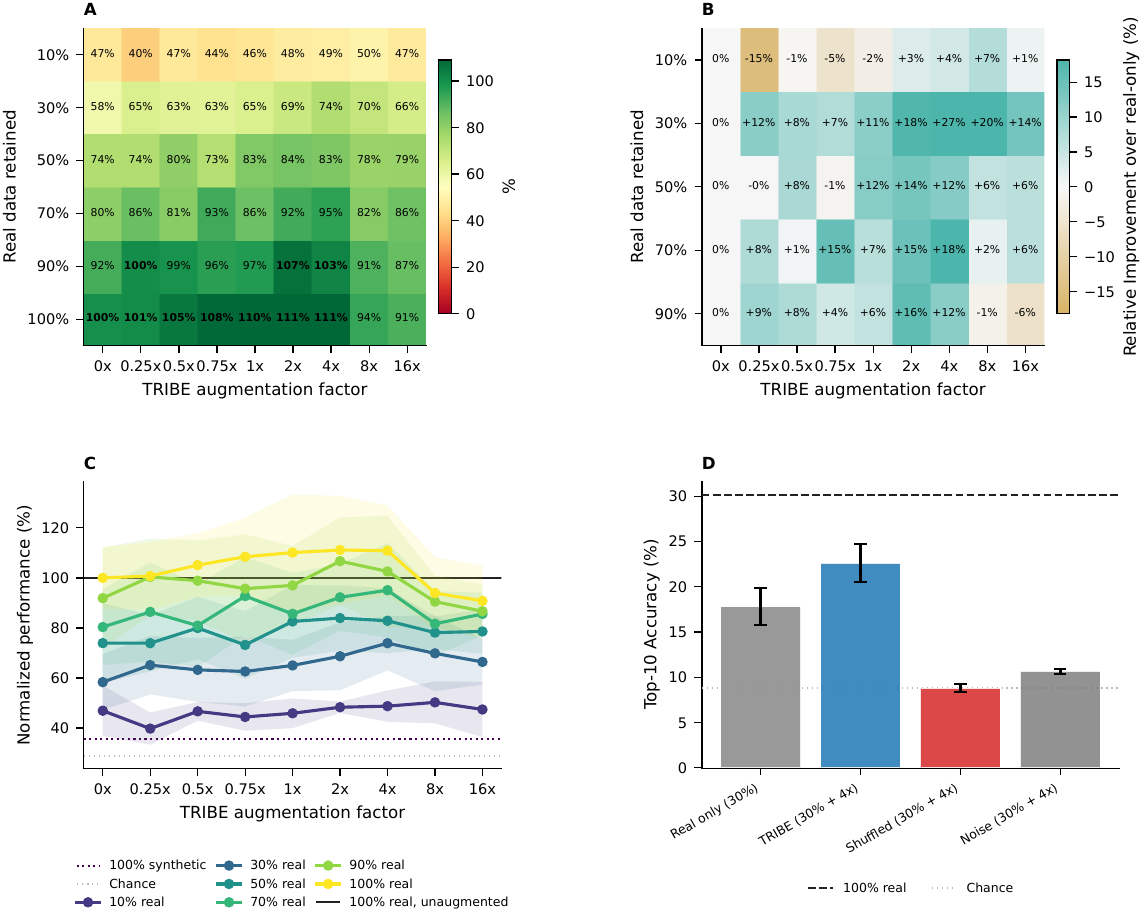}
    \caption{BOLD5000}
  \end{subfigure}
  \caption{\textbf{Operating grids for TRIBE augmentation using Deep decoders.}
  The operating grids as in Figure~\ref{fig:ridge-operating} are computed for the Deep Residual decoders.
  \textbf{A}. Top-10 retrieval accuracy normalized to the performance of a 100\% real-data-only baseline (\ie a deep model trained on the full, unaugmented training set); values above 100\% indicate that TRIBE augmentation surpasses this 100\% real-data-only baseline.
  \textbf{B}. Relative performance improvement over the matched real-only   condition ($a=0$) at the same retained-real percentage $p$\%.
  \textbf{C}. Augmentation scaling curves for each $p$\% of real-data retained, including chance and synthetic-only baselines. Shading shows SEM across all subjects (four subjects for each dataset).
  \textbf{D}. Raw Top-10 retrieval at the selected real-data fraction $p$\%, comparing TRIBE with matched real-only, Shuffled, and noise-augmentation controls; the dashed line marks the full real-only reference ($p=100\%$).
  The qualitative pattern observed for Ridge decoders remains: TRIBE helps in selected low- and medium-data regimes, but gains are less uniform than for Ridge and depend on both the dataset and the augmentation factor.
  This supports the claim that TRIBE contains useful decoding signal while emphasizing that synthetic fMRI must be calibrated to the decoder.}
  \label{fig:fmrimlp-operating}
\end{figure}

\subsection{Operating grids for Deep Decoders}
\label{sec:fmrimlp-results}

Figure~\ref{fig:fmrimlp-operating} extends the operating grids for Ridge models to deep decoders.
On both datasets, the deeper architecture can benefit from TRIBE augmentation at selected operating points, and BOLD5000 again shows stronger high-retention gains than NSD.
The effects are more modest and noisier than for Ridge decoders: performance saturates, and can start decreasing, at smaller augmentation factors, typically around $a=4$.
This may be explained by the fact that deep decoders are already strong baselines on matched real-only conditions, which makes it more difficult to push performance further with TRIBE augmentation.

Despite this noisier behavior, the same practical regimes appear in both datasets.
Around 90\% of the full real-only baseline performance can be reached with only $p=70\%$ retained real data, using $a=2$ for NSD and $a=0.75$ for BOLD5000.
Recovering the full real-only baseline requires more real data: both datasets reach it at approximately $p=90\%$ with a small amount of TRIBE augmentation ($a=0.25$).
Here too, the best relative-improvement score aligns across datasets, but at a different regime than Ridge: for Deep decoders, the largest relative gain is observed at $p=30\%$ and $a=4$ in both NSD and BOLD5000.
Finally, unlike the Ridge case in NSD, the full-synthetic reference remains above chance for Deep decoders in both datasets, suggesting that the Deep decoder can extract some usable visual signal from TRIBE predictions even without real target-dataset fMRI.
See Appendix~\ref{app:additional-metrics} for per-subject Top-10 operating grids for Deep decoders.

\subsection{Synthetic-Only Training and Controls}
\label{sec:controls-results}

The operating-grid figures include three controls.
First, the synthetic-only baseline remains above chance for Ridge models on BOLD5000 and for Deep models on both datasets (Panel C in Figures~\ref{fig:ridge-operating} and~\ref{fig:fmrimlp-operating}).
This is notable because TRIBE's visual pathway was trained on natural videos rather than static-image fMRI responses, yet its predictions still carry visual information aligned with the DINOv2-small image embedding.

Second, the Gaussian-noise control replaces the TRIBE responses with random Gaussian noise while keeping the training-set size fixed.
Its lower performance shows that the gain is not explained by simply adding samples or high-dimensional variability to the fMRI input.

Third, in the Shuffled control, we randomly reassign TRIBE responses across images while keeping the data and training protocol unchanged.
This breaks the correspondence between each image and its predicted response.
TRIBE outperforms Shuffled in all four Panel D comparisons.
More broadly, for each $a\in\{1,2,4,8\}$, the TRIBE-minus-Shuffled Top-10 difference averaged over the retained real fractions of the operating grid is positive in all eight subjects for both Ridge and Deep decoders (Appendix Table~\ref{tab:shuffled-control}).
Together, these controls show that the gain depends on image--response correspondence rather than merely adding samples or fMRI-like responses.

\subsection{Augmenting Image Reconstruction}
\label{sec:dynadiff-results}

Beyond visual-embedding retrieval, we test whether TRIBE augmentation also improves fMRI-to-image reconstruction using DynaDiff~\citep{careil2025dynadiff}, a generative decoder that reconstructs images from fMRI.
We train DynaDiff separately for the four full-protocol NSD subjects (1, 2, 5, and 7), retaining all available real fMRI ($p=100\%$) and varying the TRIBE augmentation factor $a\in\{0,0.5,1,2\}$.
The held-out test set is identical to the retrieval experiments: the common 1{,}000-image NSD test split, which contains no synthetic data.

\begin{figure}[H]
  \centering
    \includegraphics[width=0.7\linewidth]{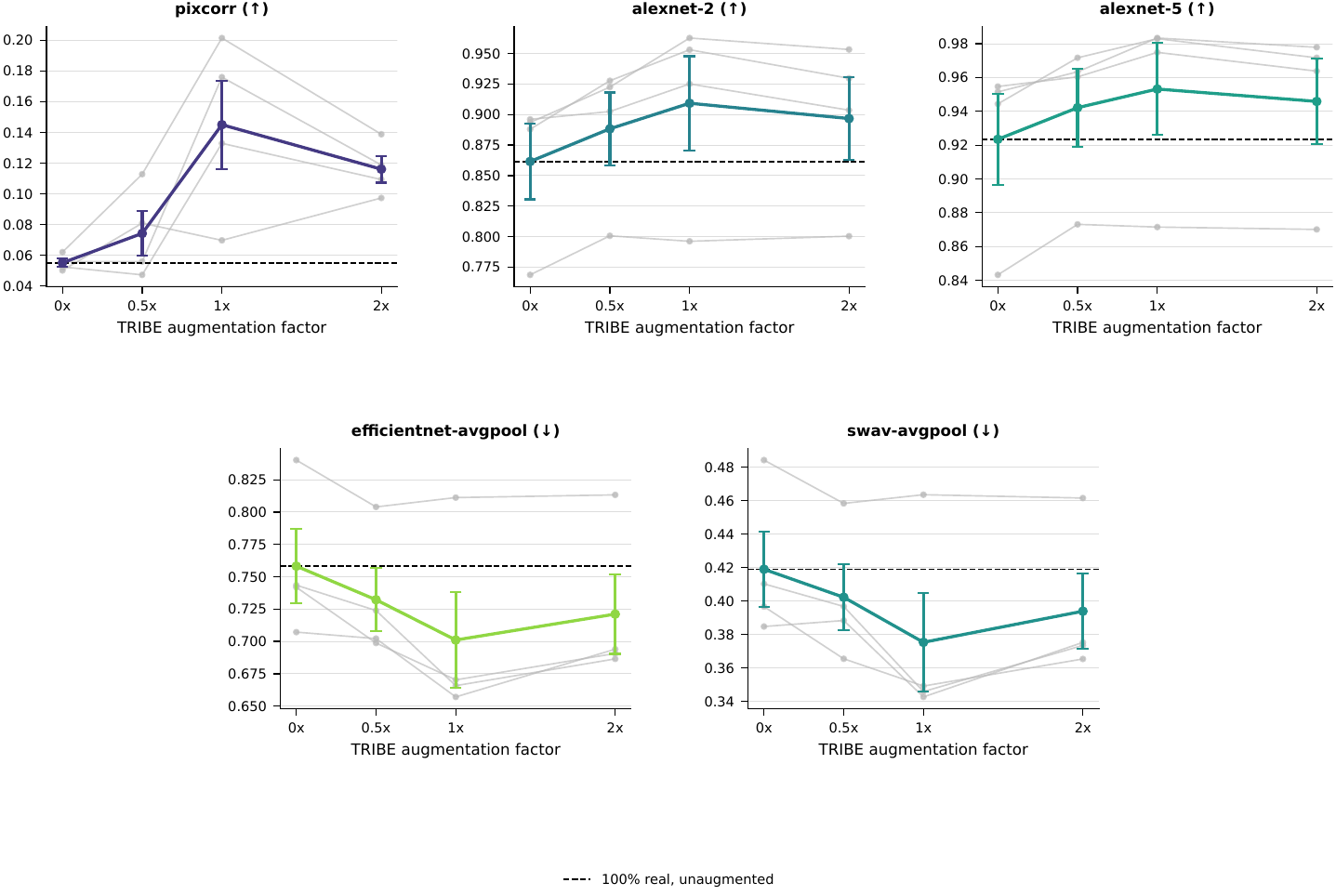}

\caption{\textbf{Image reconstruction metrics across four NSD subjects.}
Colored curves and error bars report mean $\pm$ SEM; gray curves show individual subjects.
The dashed line is the mean 100\% real, unaugmented reference.
TRIBE improves all five displayed metrics at moderate augmentation factors, with a subject-averaged optimum at $a=1$.}
  \label{fig:dynadiff}
\end{figure}

Figure~\ref{fig:dynadiff} reports five reconstruction metrics as mean $\pm$ SEM across subjects, together with the individual-subject trajectories.
At the subject-averaged optimum $a=1$, PixCorr increases from $0.055$ to $0.145$, AlexNet-2/5 increase from $0.862/0.923$ to $0.909/0.953$, and EfficientNet/SwAV distances decrease from $0.758/0.419$ to $0.701/0.375$.
That is, all five metrics improve for all four subjects at $a=1$.
At $a=2$, the subject-averaged metrics regress toward the real-only baseline while still improving on it.
Across the full eight-metric suite, seven metrics improve for all four subjects at $a=1$, while SSIM changes little on average and improves for two of the four subjects (Figure~\ref{fig:dynadiff-all-metrics} in Appendix~\ref{app:dynadiff}).
Given the four-subject sample, we interpret these results as consistent cross-subject evidence rather than population-level inference.

\section{Discussion}
\label{sec:discussion}

\paragraph{Synthetic fMRI augmentation only helps in specific data-regimes.}
Our results show that TRIBE~v2\citep{dascoli2026tribe} augmentation can improve Top-10 brain-to-image retrieval accuracy by up to 68\% relative to matched real-only training when fMRI data are scarce. %
As expected, such gains depend both on the amount of available data and on the proportion of synthetic data used; beyond a certain point, decoding performance saturates and can even deteriorate as more synthetic data are added. %
Our operating grids characterize these benefits retrospectively rather than predict them a priori.
In practice, the synthetic-to-real ratio should be selected on validation data for each dataset and decoder.

\paragraph{Out-of-distribution pretraining.}
It is quite remarkable that this approach works at all: TRIBE~v2\citep{dascoli2026tribe} is neither trained on brain responses to static images (or static image embedding) nor is it trained for decoding. Its visual pathway is trained on brain responses to long movies, jointly with audio and language. Here, we use repeated-frame 3\,s clips as the reference protocol; duration and motion sensitivity checks yield similar augmentation performance (Appendix Table~\ref{tab:videoization-control}). Consequently, the present results suggest that TRIBE captures patterns of visual responses that transfer from naturalistic multimodal stimulation to a more controlled rapid-serial-visual-presentation of images. At the same time, this mismatch may also explain why augmentation benefits saturate and why the optimal synthetic-to-real ratio must be calibrated. %

\paragraph{Why can augmentation exceed the full-real reference?}
Some configurations exceed the 100\% real reference.
This should not be read as synthetic fMRI containing more subject-specific information than real fMRI.
A more plausible explanation is that TRIBE~v2 adds stimulus diversity and population-level visual response structure that regularize the fMRI-to-embedding mapping, especially when the real training set is small or idiosyncratic.
Because the augmentation pool excludes all images seen by the target subject and evaluation is always on real fMRI, these gains are not caused by overlap with evaluation stimuli or by testing on synthetic data.

\paragraph{Subject-agnostic augmentation has limits.}
Here, TRIBE~v2 predicts an average-subject brain response (i.e. there is no target-subject adaptation).
This makes the augmentation broadly deployable, but it also means synthetic responses lack individual-specific anatomy and response idiosyncrasies.
The synthetic-only reference captures this limitation: above-chance decoding is possible, but the strongest results still require real fMRI from the target subjects.
Future work could combine the operating grid protocol with subject-specific TRIBE adaptation, learned synthetic-data weighting, or multi-subject decoder pretraining~\citep{scotti2024mindeye2}.

\paragraph{Limitations.}
\textit{First}, although we replicate our findings on two datasets, the latter have a small number of subjects. Extension to additional datasets, including other modalities thus remains to be further evaluated. Given TRIBE~v2 is multimodal, our approach should not require major changes.
\textit{Second}, the primary task focuses on image retrieval from a DINOv2-small embedding~\citep{oquab2023dinov2}. Image reconstruction is evaluated on the four full-protocol NSD subjects; the four partial-protocol subjects are excluded because they do not share the common 1{,}000-image test set. This choice of tasks was also constrained by the fact that most available fMRI-to-image models (e.g. \citet{scotti2024mindeye,scotti2024mindeye2}) are based on time-independent fMRI beta maps, while TRIBE~v2 generates dynamic fMRI signals. It will be important to extend the present tests to other fMRI-to-image models and datasets, \eg by deriving beta maps from TRIBE~v2 with preprocessing matched to NSD~\citep{allen2022massive} and BOLD5000~\citep{chang2019bold5000} betas.

\textit{Finally}, the published training-data descriptions for DINOv2 and V-JEPA~2 do not list COCO as a source, but their web-scale corpora prevent us from exhaustively ruling out prior exposure to COCO images or perceptual near-duplicates.
Although we exclude exact target-study stimuli from the augmentation pools, we do not claim complete exclusion of all possible visual-backbone pretraining exposure.

\paragraph{Broader impact.}
If reliable, model-based fMRI augmentation could make image-decoding research less dependent on very large scan-time budgets and therefore easier to study across laboratories and populations.
At the same time, the results should not be interpreted as enabling decoding without any real data: the best performance still relies on subject-specific real fMRI responses.

\section{Conclusion}
\label{sec:conclusion}
Synthetic fMRI augmentation could mark a shift toward more accessible neuroimaging by reducing the reliance on massive scan-time budgets. While TRIBE v2 demonstrates that naturalistic, multimodal pretraining can effectively regularize decoding in data-scarce regimes, the future of the field lies in balancing these general population-level priors with subject-specific nuances. Ultimately, calibrating the synergy between synthetic diversity and real neural signals will be essential for building efficient decoders of brain activity.

\section*{Acknowledgments and Disclosure of Funding}
Collection of the NSD dataset was supported by NSF IIS-1822683 and NSF IIS-1822929.

{
\small

\bibliographystyle{assets/plainnat}
\bibliography{references}

}

\clearpage
\beginappendix

\section{MindEye-Style MLP Decoder Architecture}
\label{app:fmrimlp-architecture}

The deep decoder used in the operating grids follows the residual MLP architecture introduced by MindEye~\citep{scotti2024mindeye}, with the parameterization described below.
The input is a one-TR fsaverage5 response with 20{,}484 vertices, and the output is a 384-dimensional DINOv2-small image embedding.
Table~\ref{tab:fmrimlp-architecture} reports the architecture used in our experiments: hidden width 553, two residual blocks, an input projection from fsaverage5 vertices to the hidden width, and a 384-dimensional output head.
We optimize this model with AdamW (learning rate $5\times 10^{-4}$, weight decay 0.01) and a OneCycle learning-rate schedule with maximum learning rate $10^{-3}$.
Training runs for up to 40 epochs with early stopping on validation loss (patience 10) and batch size 128. Training and evaluation were run on a single NVIDIA V100 GPU with 16~GB of memory; the longest training runs took approximately 2~hours.

\begin{table}[H]
  \centering
  \small
  \caption{\textbf{Residual MLP decoder architecture used for deep-decoder experiments.}
  Parameter counts are for one per-subject decoder. All parameters are trainable.}
  \label{tab:fmrimlp-architecture}
  \begin{tabular}{llr}
    \toprule
    Component & Operation & Parameters \\
    \midrule
    Vertex-to-hidden map & Linear map, $20{,}484 \rightarrow 553$, no bias & 11{,}327{,}652 \\
    Hidden mixing layer & Linear map within the hidden space, $553 \rightarrow 553$ & 306{,}362 \\
    Post-TR block & LayerNorm(553), GELU, dropout $p=0.5$ & 1{,}106 \\
    Residual MLP blocks & $2\times$ [Linear $553 \rightarrow 553$, LayerNorm, GELU, dropout $p=0.15$] & 614{,}936 \\
    Temporal aggregation & Linear aggregation over the single retained TR & 2 \\
    Embedding readout & Linear $553 \rightarrow 384$ & 212{,}736 \\
    Output head & Linear projection $384 \rightarrow 384$, dropout $p=0$ & 147{,}840 \\
    \midrule
    Total & & 12{,}610{,}634 \\
    \bottomrule
  \end{tabular}
\end{table}

\section{Additional Operating Grids}
\label{app:additional-metrics}

\begin{figure}[H]
  \centering
  \begin{subfigure}[t]{0.49\linewidth}
    \centering
    \includegraphics[width=\linewidth]{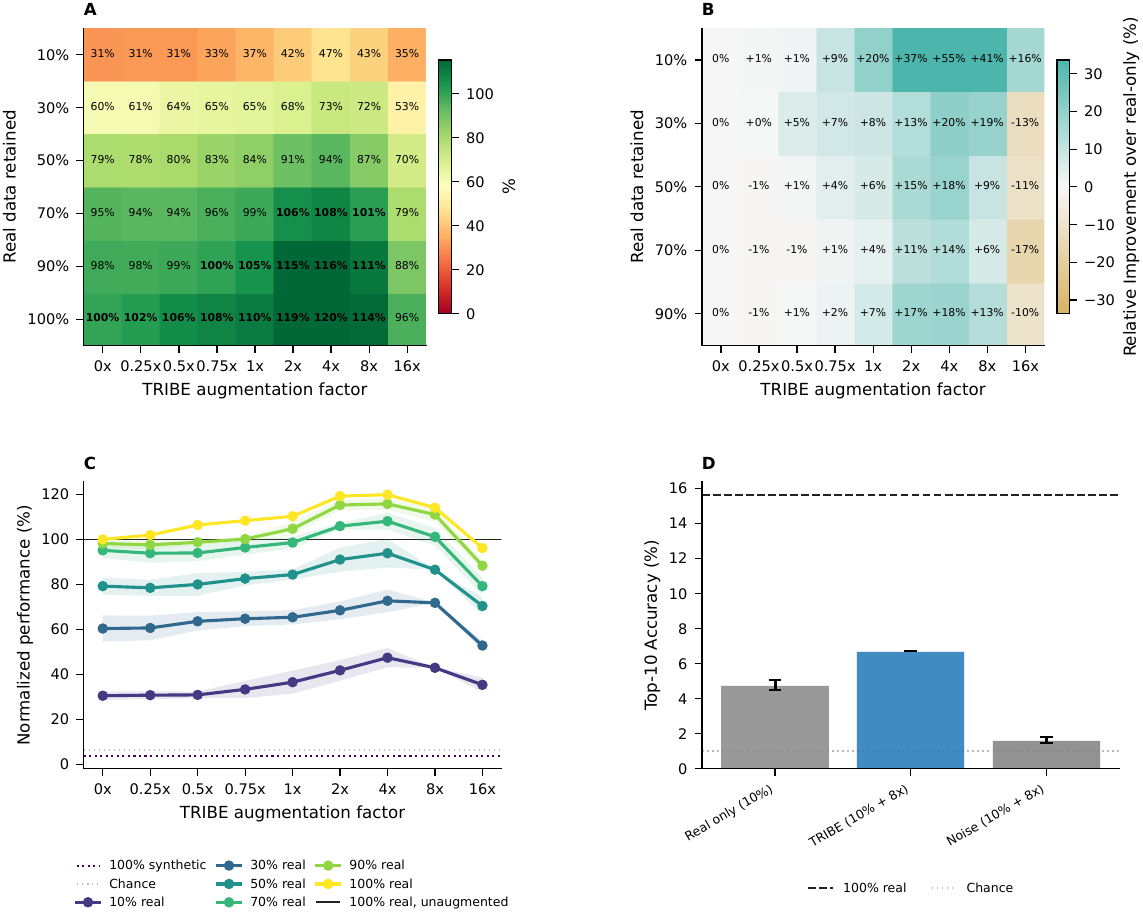}
    \caption{Subject 1.}
  \end{subfigure}
  \hfill
  \begin{subfigure}[t]{0.49\linewidth}
    \centering
    \includegraphics[width=\linewidth]{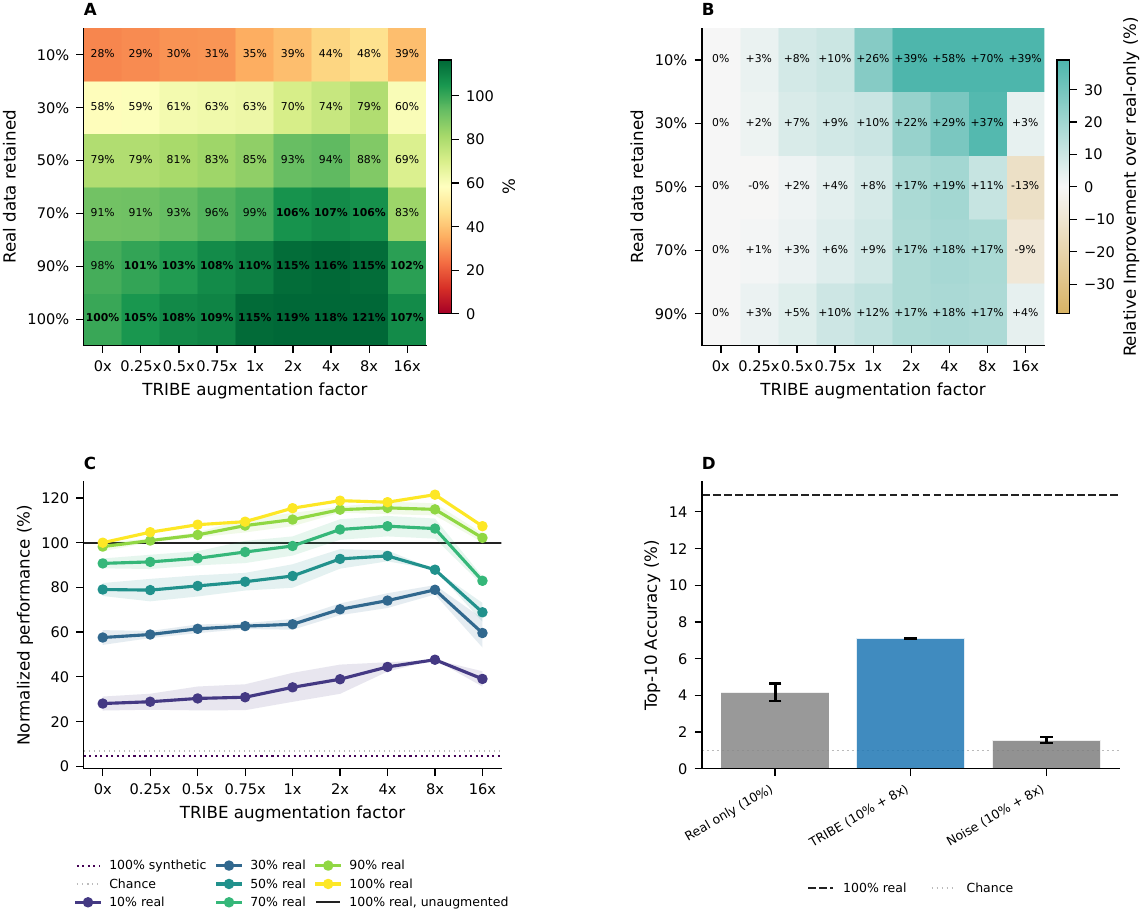}
    \caption{Subject 2.}
  \end{subfigure}
  \begin{subfigure}[t]{0.49\linewidth}
    \centering
    \includegraphics[width=\linewidth]{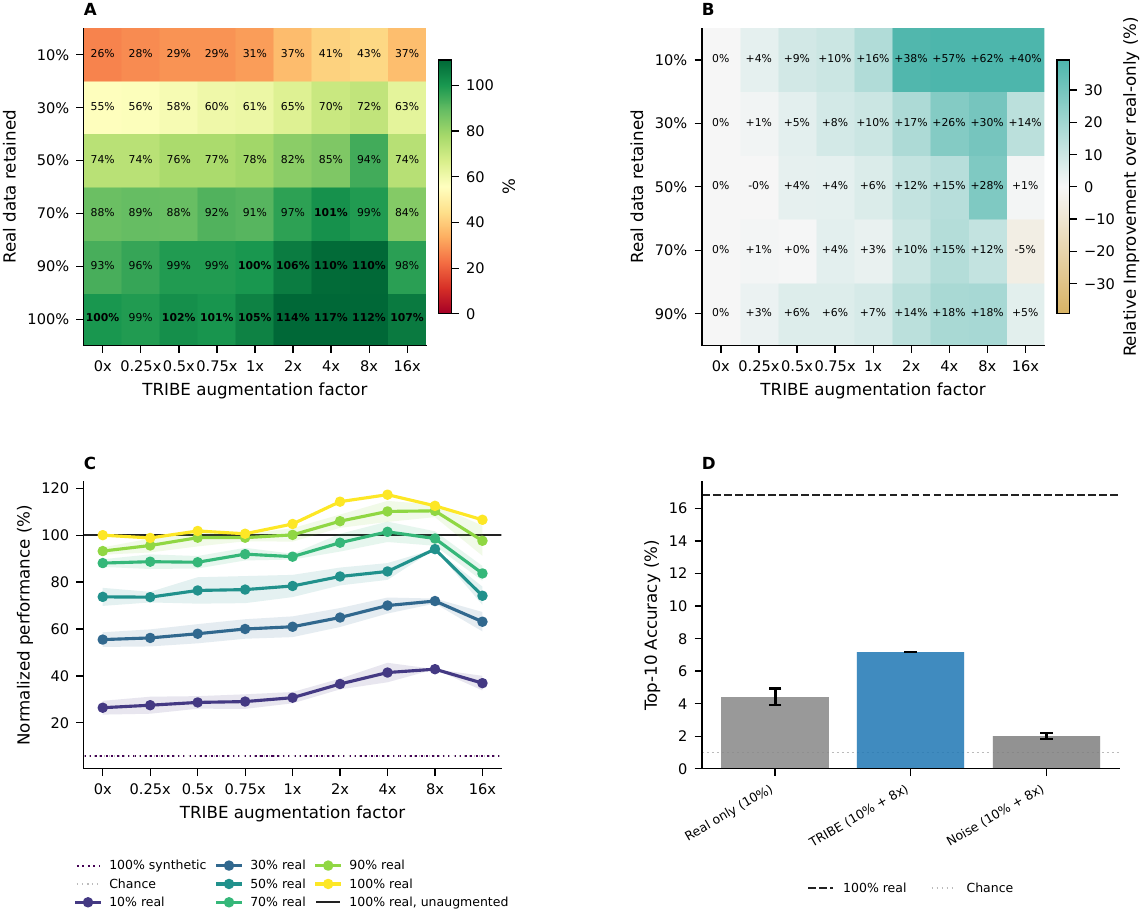}
    \caption{Subject 5.}
  \end{subfigure}
  \hfill
  \begin{subfigure}[t]{0.49\linewidth}
    \centering
    \includegraphics[width=\linewidth]{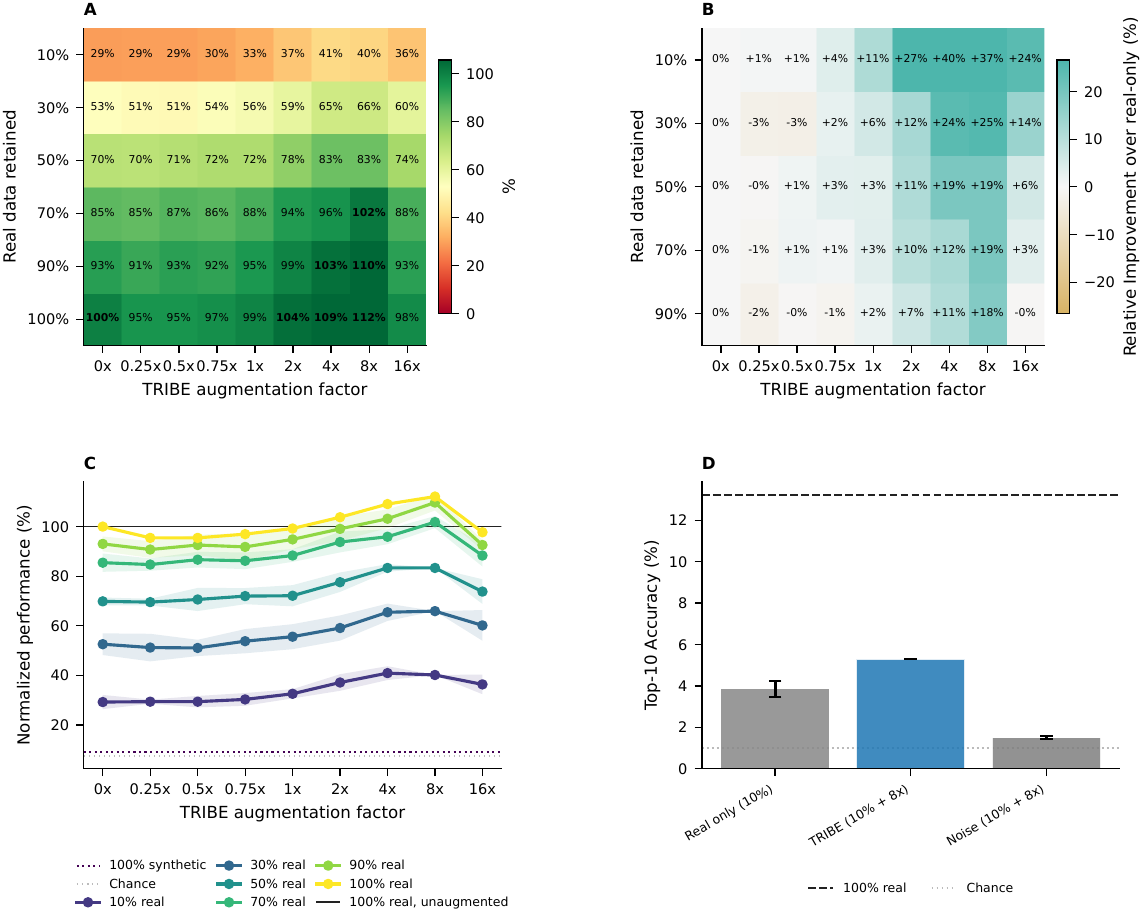}
    \caption{Subject 7.}
  \end{subfigure}
  \caption{\textbf{Per-subject Top-10 operating grids for Ridge decoders on NSD.}
  Each subfigure reports one subject rather than averaging across subjects, using the same normalization, retained-real fractions, and augmentation factors as Figure~\ref{fig:ridge-operating}.}
  \label{fig:app-ridge-allen-subjects}
\end{figure}

\begin{figure}[H]
  \centering
  \begin{subfigure}[t]{0.49\linewidth}
    \centering
    \includegraphics[width=\linewidth]{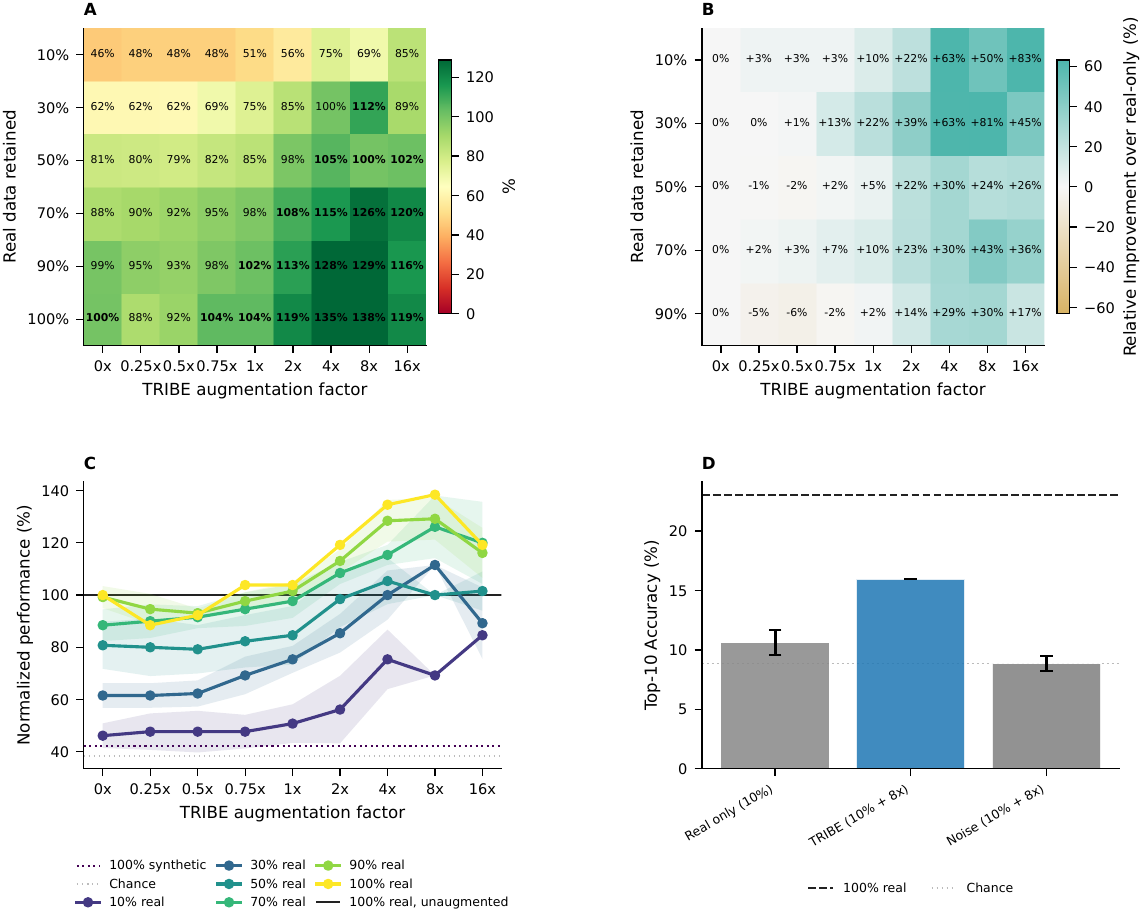}
    \caption{Subject 1.}
  \end{subfigure}
  \hfill
  \begin{subfigure}[t]{0.49\linewidth}
    \centering
    \includegraphics[width=\linewidth]{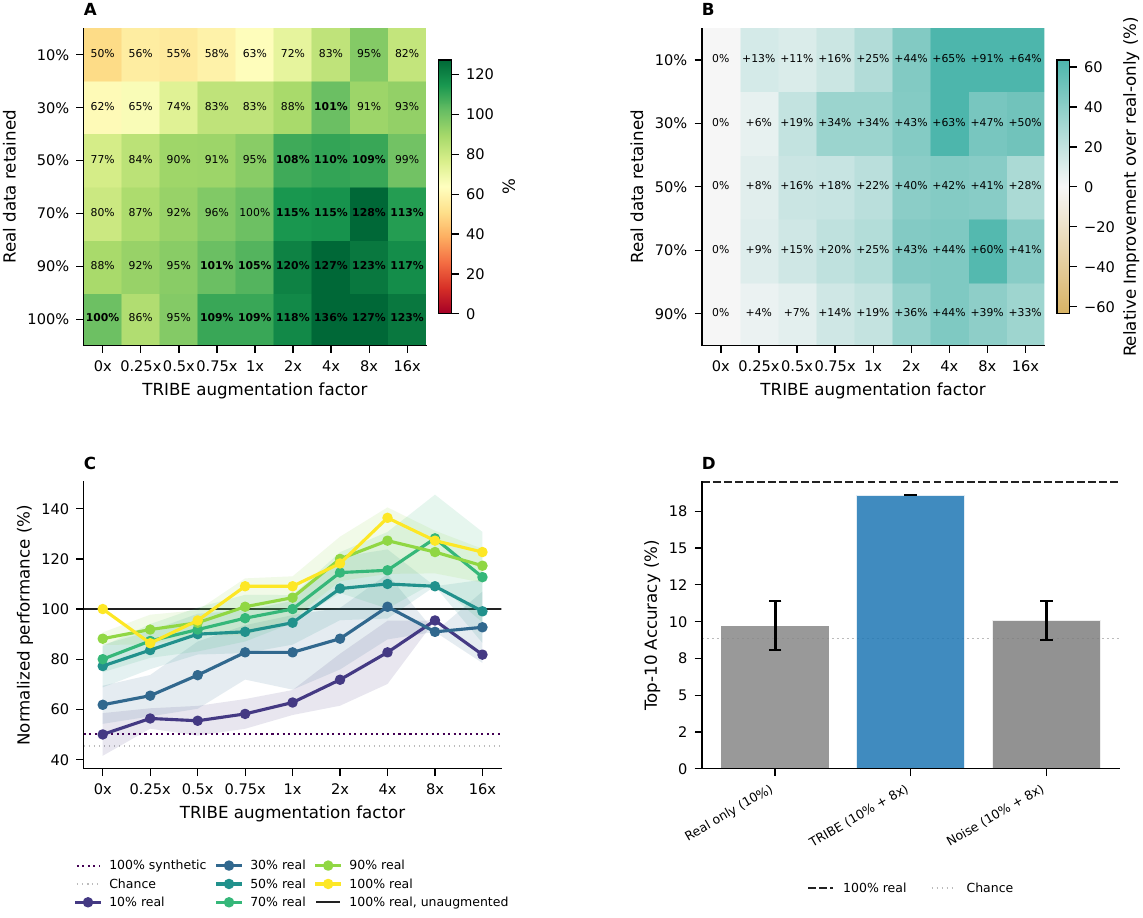}
    \caption{Subject 2.}
  \end{subfigure}
  \begin{subfigure}[t]{0.49\linewidth}
    \centering
    \includegraphics[width=\linewidth]{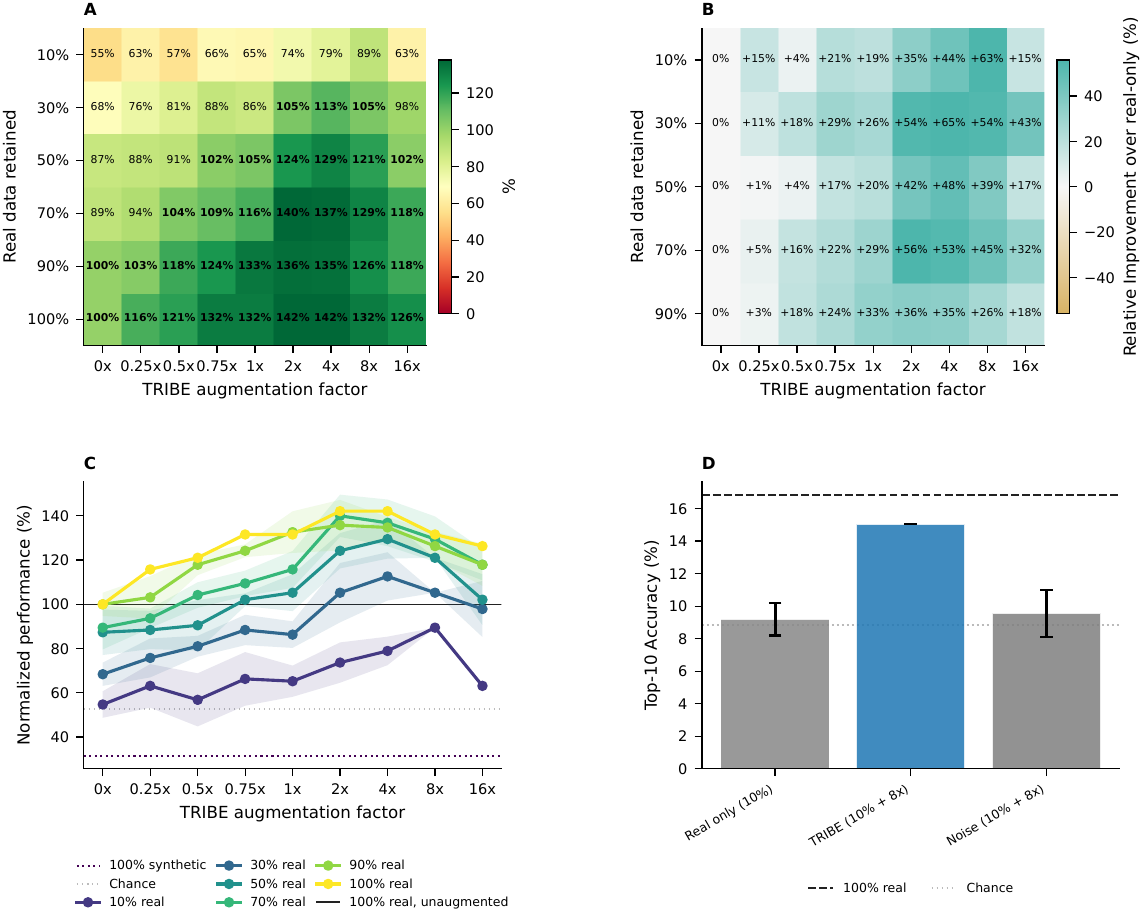}
    \caption{Subject 3.}
  \end{subfigure}
  \hfill
  \begin{subfigure}[t]{0.49\linewidth}
    \centering
    \includegraphics[width=\linewidth]{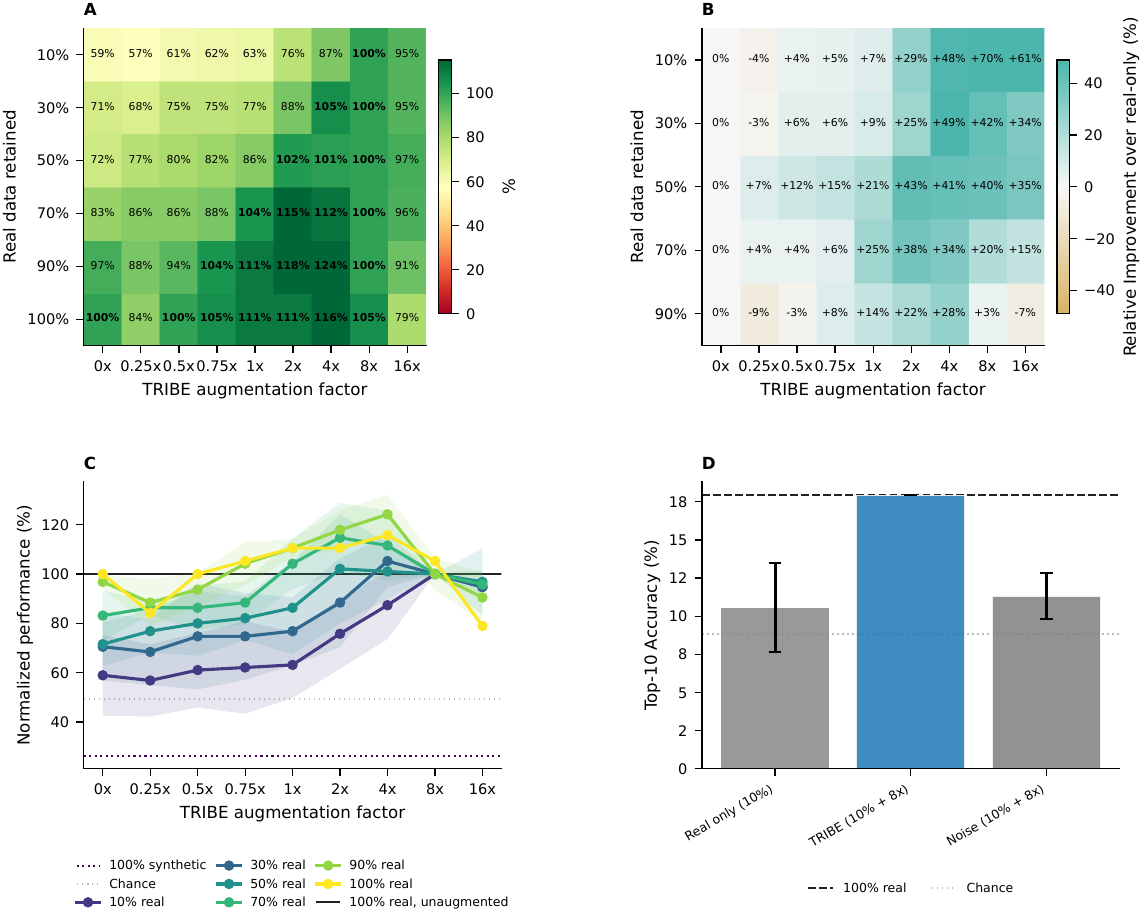}
    \caption{Subject 4.}
  \end{subfigure}
  \caption{\textbf{Per-subject Top-10 operating grids for Ridge decoders on BOLD5000.}
  Each subfigure reports one subject rather than averaging across subjects, using the same normalization, retained-real fractions, and augmentation factors as Figure~\ref{fig:ridge-operating}.}
  \label{fig:app-ridge-chang-subjects}
\end{figure}

\begin{figure}[H]
  \centering
  \begin{subfigure}[t]{0.49\linewidth}
    \centering
    \includegraphics[width=\linewidth]{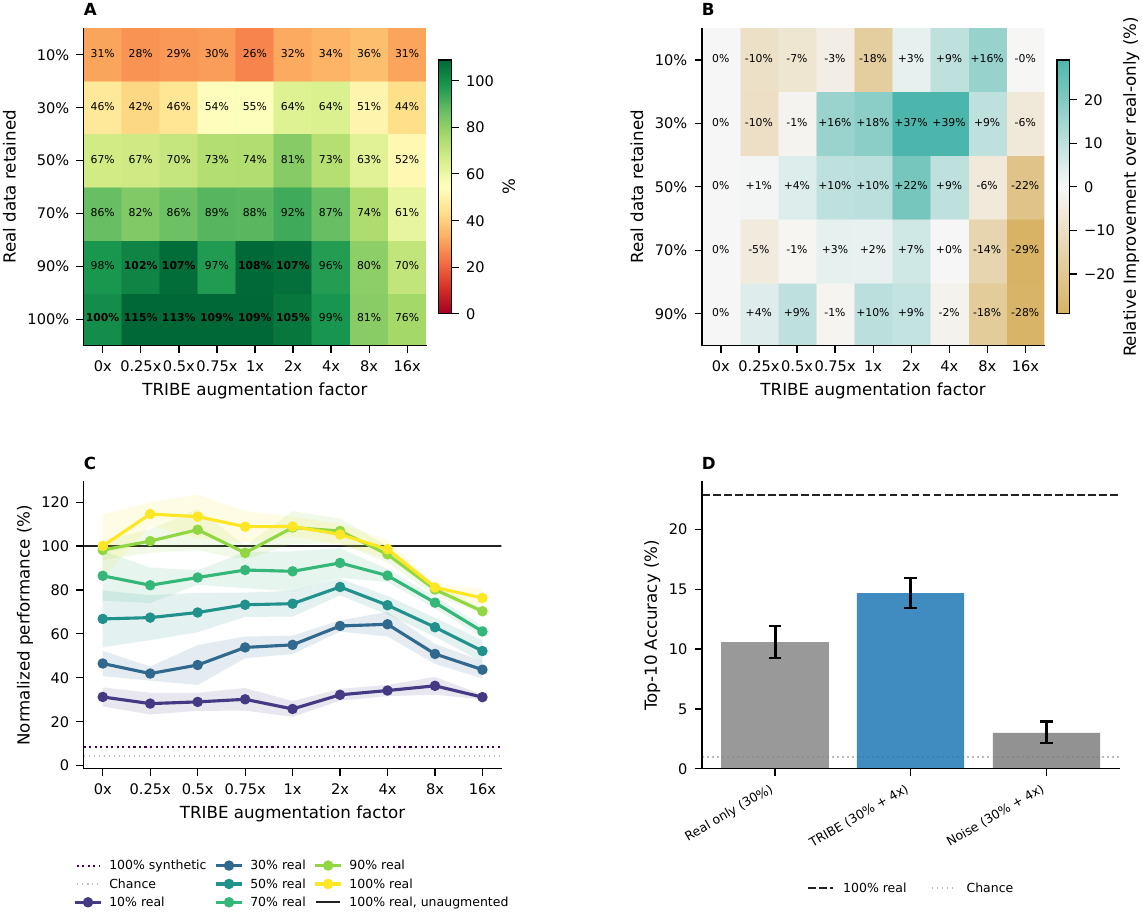}
    \caption{Subject 1.}
  \end{subfigure}
  \hfill
  \begin{subfigure}[t]{0.49\linewidth}
    \centering
    \includegraphics[width=\linewidth]{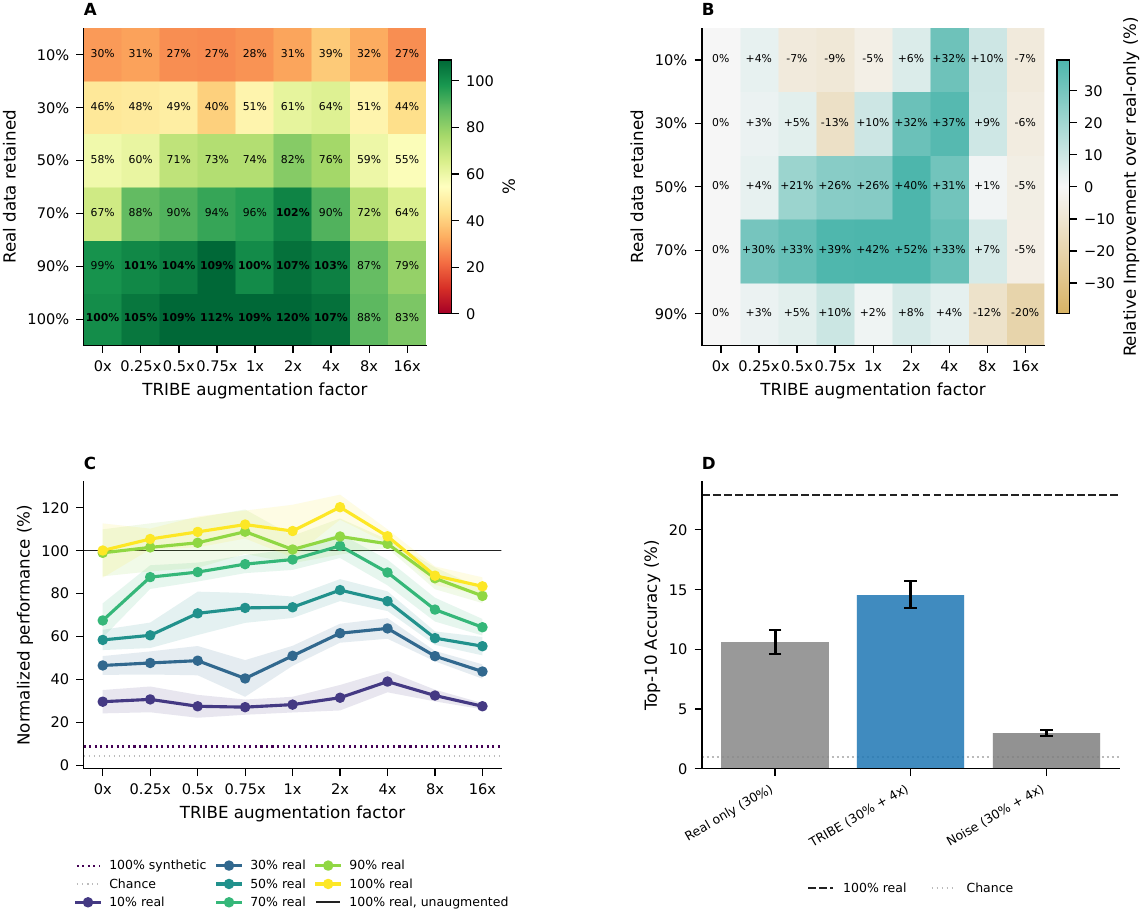}
    \caption{Subject 2.}
  \end{subfigure}
  \begin{subfigure}[t]{0.49\linewidth}
    \centering
    \includegraphics[width=\linewidth]{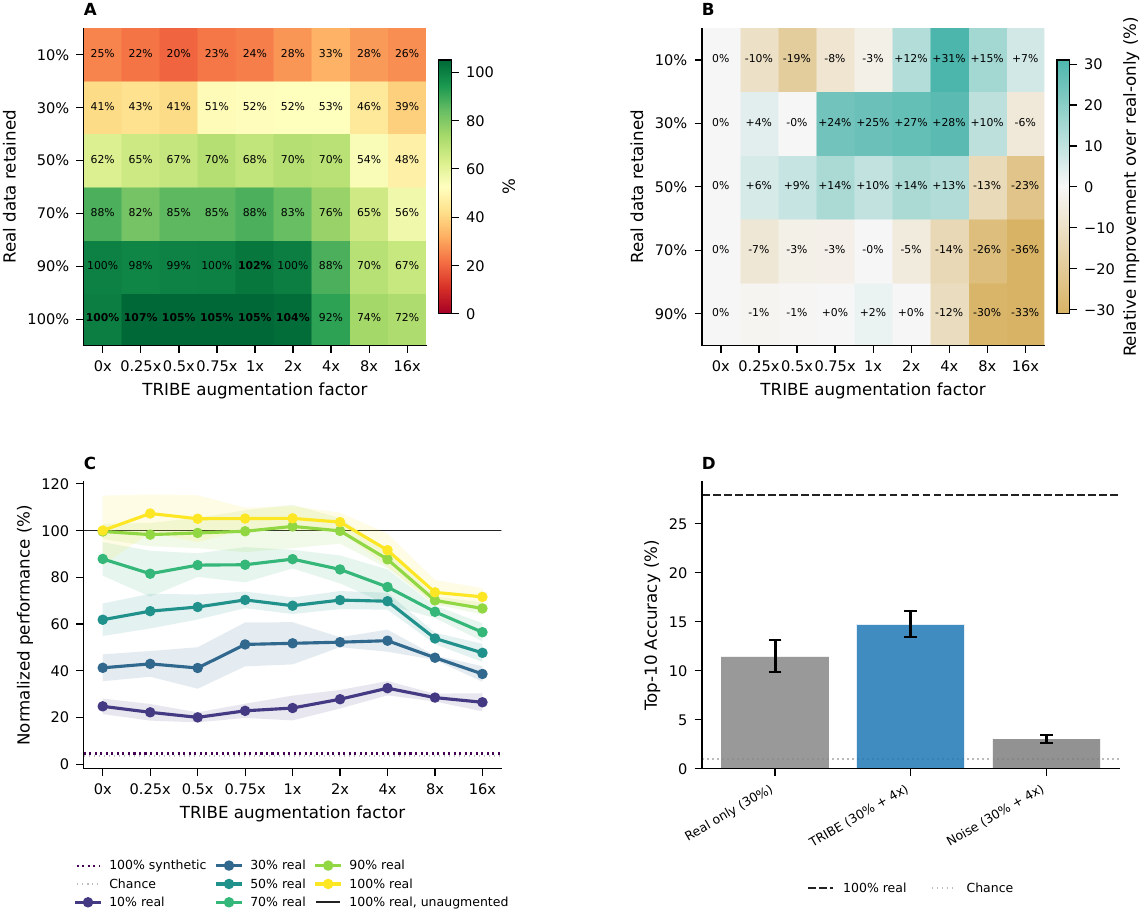}
    \caption{Subject 5.}
  \end{subfigure}
  \hfill
  \begin{subfigure}[t]{0.49\linewidth}
    \centering
    \includegraphics[width=\linewidth]{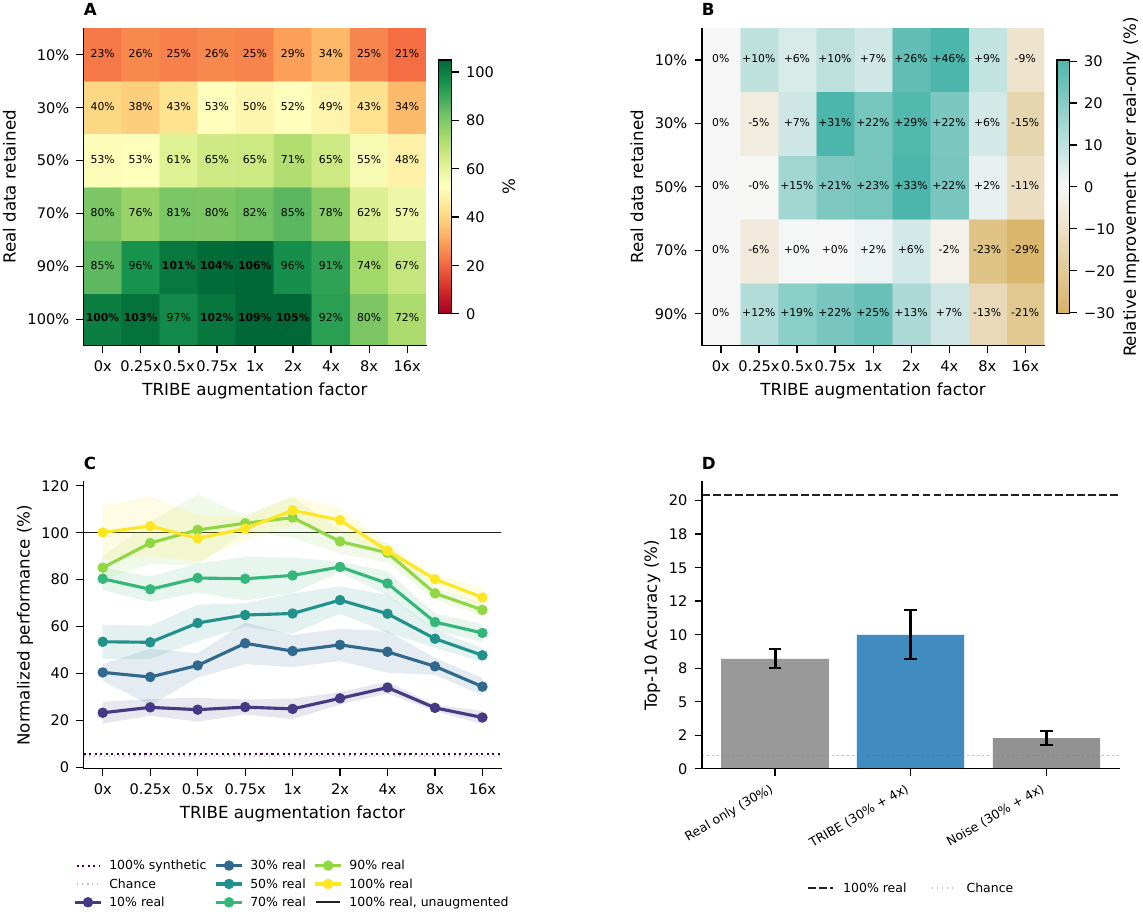}
    \caption{Subject 7.}
  \end{subfigure}
  \caption{\textbf{Per-subject Top-10 operating grids for Deep decoders on NSD.}
  Each subfigure reports one subject rather than averaging across subjects, using the same normalization, retained-real fractions, and augmentation factors as Figure~\ref{fig:fmrimlp-operating}.}
  \label{fig:app-fmrimlp-allen-subjects}
\end{figure}

\begin{figure}[H]
  \centering
  \begin{subfigure}[t]{0.49\linewidth}
    \centering
    \includegraphics[width=\linewidth]{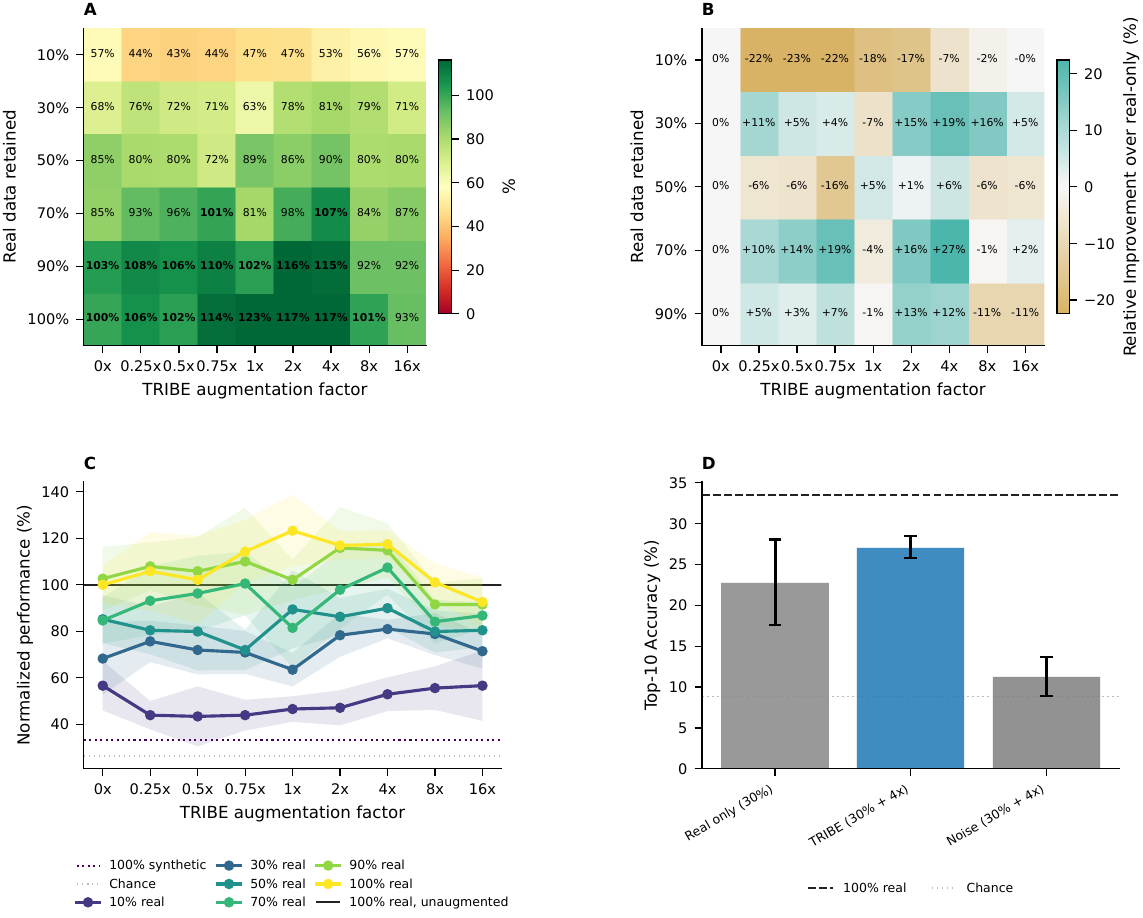}
    \caption{Subject 1.}
  \end{subfigure}
  \hfill
  \begin{subfigure}[t]{0.49\linewidth}
    \centering
    \includegraphics[width=\linewidth]{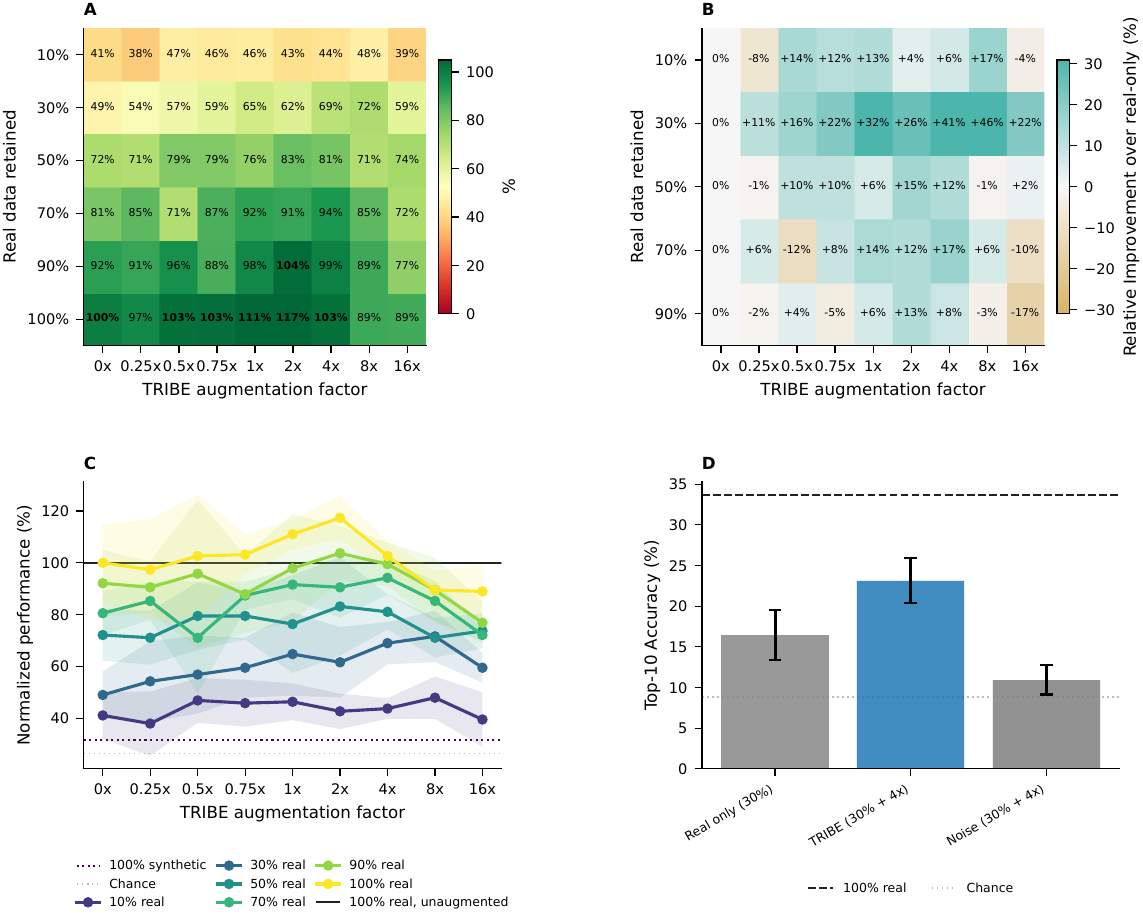}
    \caption{Subject 2.}
  \end{subfigure}
  \begin{subfigure}[t]{0.49\linewidth}
    \centering
    \includegraphics[width=\linewidth]{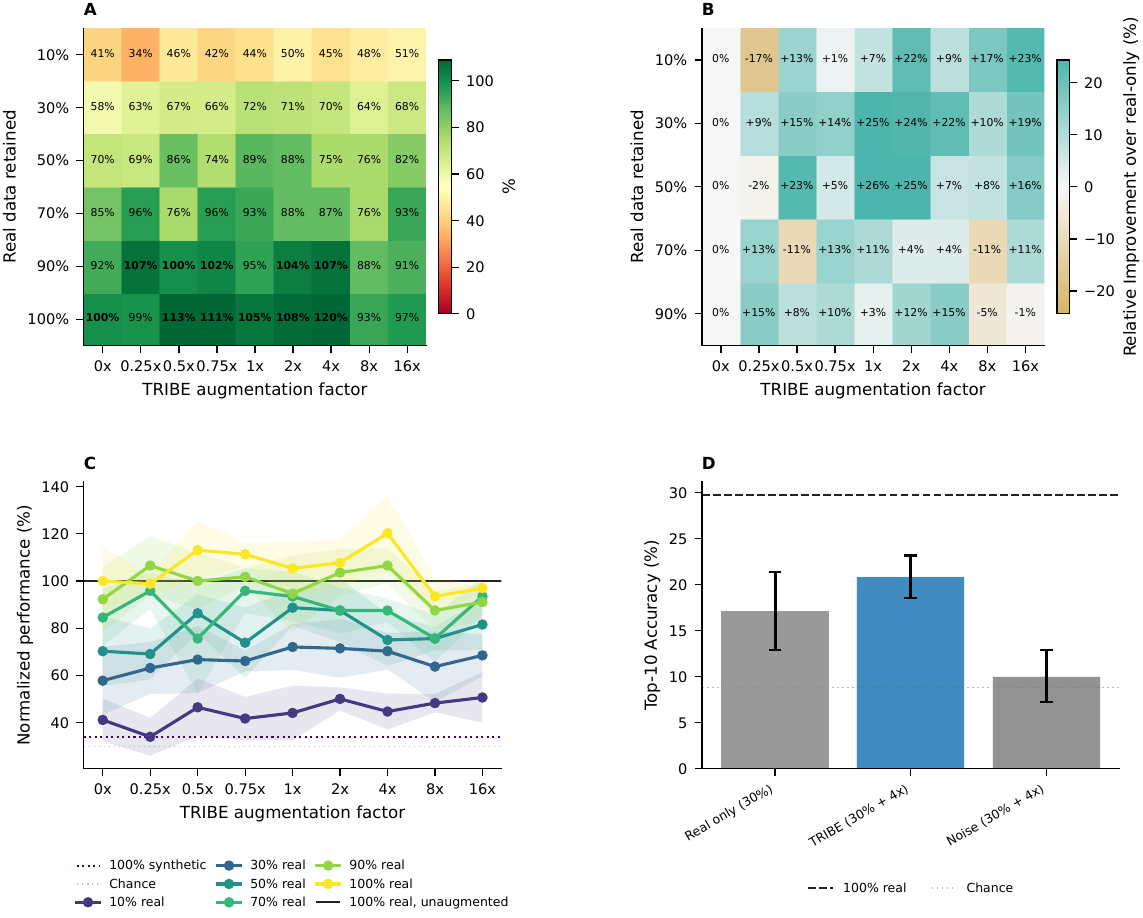}
    \caption{Subject 3.}
  \end{subfigure}
  \hfill
  \begin{subfigure}[t]{0.49\linewidth}
    \centering
    \includegraphics[width=\linewidth]{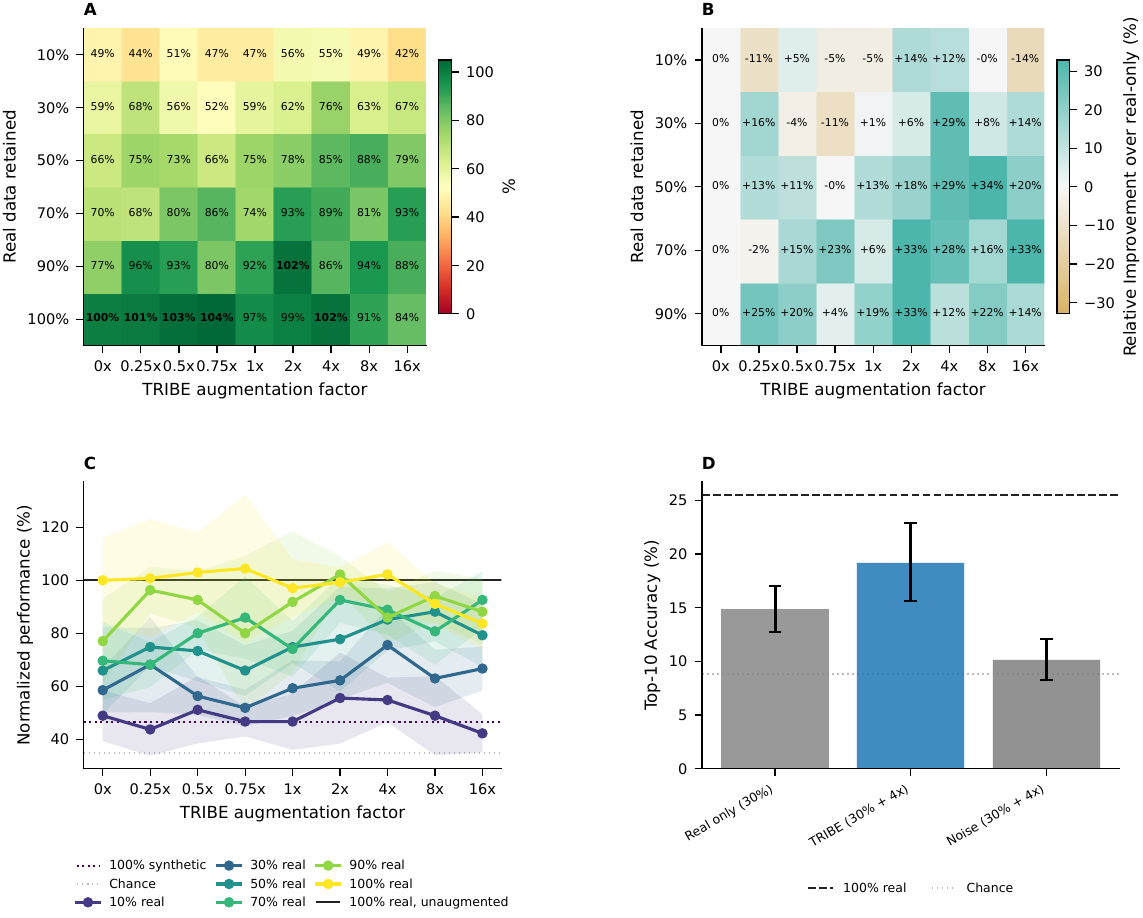}
    \caption{Subject 4.}
  \end{subfigure}
  \caption{\textbf{Per-subject Top-10 operating grids for Deep decoders on BOLD5000.}
  Each subfigure reports one subject rather than averaging across subjects, using the same normalization, retained-real fractions, and augmentation factors as Figure~\ref{fig:fmrimlp-operating}.}
  \label{fig:app-fmrimlp-chang-subjects}
\end{figure}

Figures~\ref{fig:app-ridge-allen-subjects}--\ref{fig:app-fmrimlp-chang-subjects} show the per-subject Top-10 operating grids underlying the subject-averaged results in the main text.
These grids use the same normalization, retained-real percentages, augmentation factors, and panel layout as Figures~\ref{fig:ridge-operating} and~\ref{fig:fmrimlp-operating}, but avoid averaging across subjects.
They make visible the heterogeneity of the augmentation effect: some subjects benefit from TRIBE across a broad range of low-data regimes, whereas others show narrower optima or saturation at smaller augmentation factors.
This subject-level variability is expected due to the subject-agnostic nature of TRIBE~v2 synthetic responses and motivates reporting the operating regime rather than a single global augmentation factor.

\begin{figure}[H]
  \centering
  \begin{subfigure}[t]{0.49\linewidth}
    \centering
    \includegraphics[width=\linewidth]{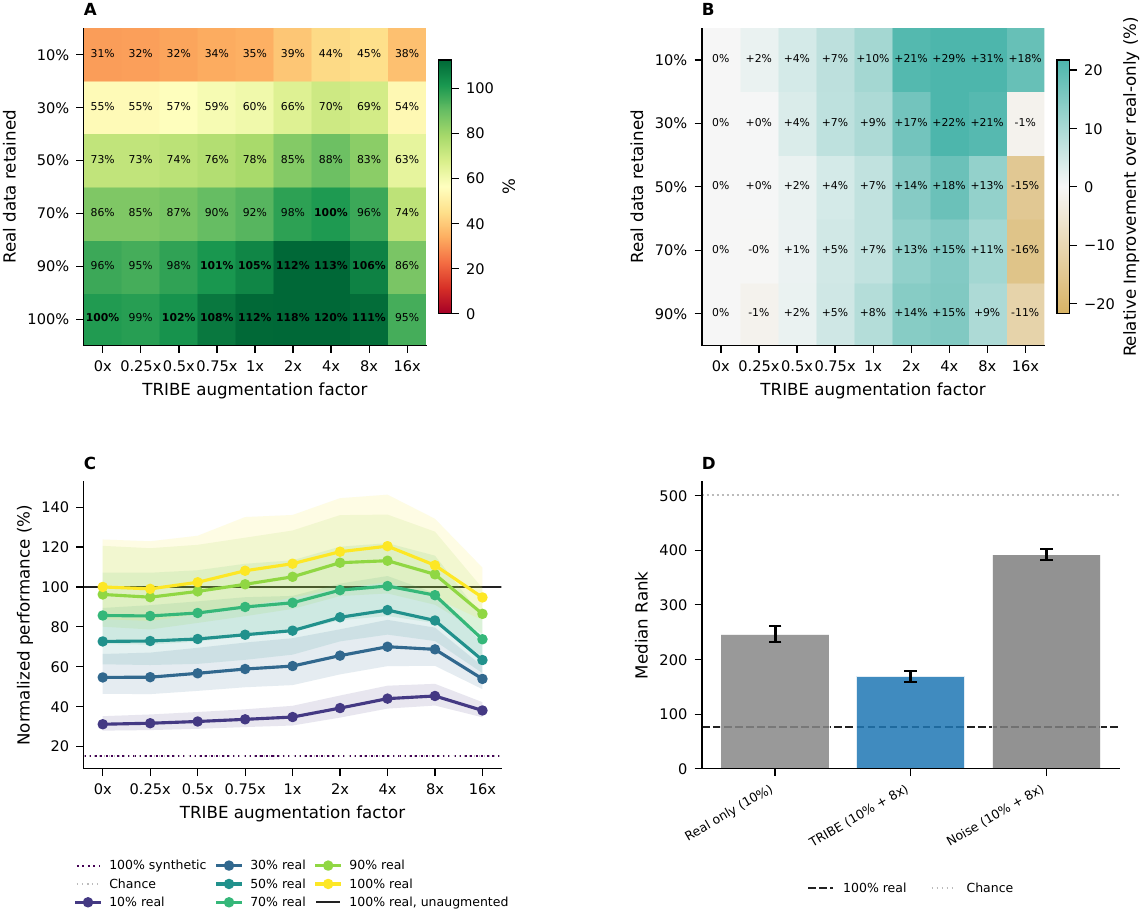}
    \caption{NSD.}
  \end{subfigure}
  \hfill
  \begin{subfigure}[t]{0.49\linewidth}
    \centering
    \includegraphics[width=\linewidth]{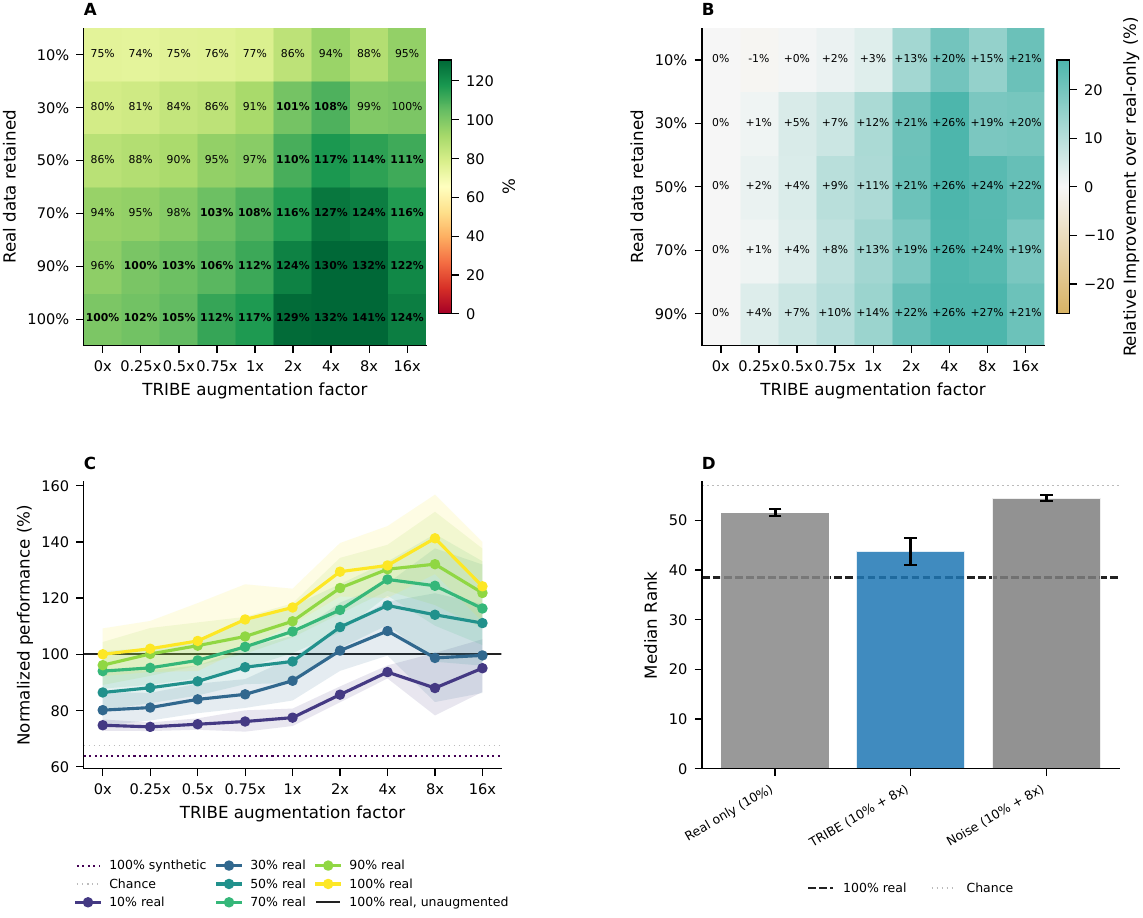}
    \caption{BOLD5000.}
  \end{subfigure}
  \caption{\textbf{Median-rank operating grids for Ridge decoders.}
  Panel D reports raw median rank (lower is better); Panels A--C are normalized so that larger values indicate better performance, matching the Top-10 figures in the main text.}
  \label{fig:app-median-rank}
\end{figure}

\subsection{Top-1 and Top-5 Retrieval}
\label{app:top1-top5}

Figures~\ref{fig:app-top1-ridge}--\ref{fig:app-top5-fmrimlp} report complete Top-1 and Top-5 operating maps.
Panel A reports raw retrieval accuracy, Panel B normalizes accuracy to the corresponding 100\% real-only baseline, Panel C shows raw scaling curves with SEM across subjects, and Panel D reports relative improvement over the matched real-only condition.
Chance accuracy is 0.1\%/0.5\% for Top-1/Top-5 on NSD and approximately 0.9\%/4.5\% on BOLD5000.

The qualitative pattern is consistent with Top-10 retrieval.
Top-5 closely follows the main results, whereas Top-1 is stricter and noisier, particularly for Deep decoders on BOLD5000.
Across metrics, gains are strongest in low- and medium-data regimes overall, although Ridge decoders on BOLD5000 also benefit at high retention.
Deep decoders show less uniform gains, and performance often saturates or declines at high augmentation factors.

\newcommand{\additionalaveragemap}[4]{%
\begin{figure}[H]
  \centering
  \begin{subfigure}[t]{0.49\linewidth}
    \centering
    \includegraphics[width=\linewidth]{figures/allen/fig_additional_#3_#4.pdf}
    \caption{NSD.}
  \end{subfigure}
  \hfill
  \begin{subfigure}[t]{0.49\linewidth}
    \centering
    \includegraphics[width=\linewidth]{figures/chang/fig_additional_#3_#4.pdf}
    \caption{BOLD5000.}
  \end{subfigure}
  \caption{\textbf{Subject-averaged #1 operating maps for #2 decoders.}
  Panels A--D report raw accuracy, normalized accuracy, raw scaling curves with SEM across subjects, and relative improvement over matched real-only, respectively.}
  \label{fig:app-#3-#4}
\end{figure}
}

\additionalaveragemap{Top-1}{Ridge}{top1}{ridge}
\additionalaveragemap{Top-1}{Deep}{top1}{fmrimlp}
\additionalaveragemap{Top-5}{Ridge}{top5}{ridge}
\additionalaveragemap{Top-5}{Deep}{top5}{fmrimlp}

\newcommand{\additionalnsdsubjectmaps}[4]{%
\begin{figure}[H]
  \centering
  \begin{subfigure}[t]{0.49\linewidth}
    \centering
    \includegraphics[width=\linewidth]{figures/allen/fig_additional_#3_#4_subject-1.pdf}
    \caption{Subject 1.}
  \end{subfigure}
  \hfill
  \begin{subfigure}[t]{0.49\linewidth}
    \centering
    \includegraphics[width=\linewidth]{figures/allen/fig_additional_#3_#4_subject-2.pdf}
    \caption{Subject 2.}
  \end{subfigure}
  \begin{subfigure}[t]{0.49\linewidth}
    \centering
    \includegraphics[width=\linewidth]{figures/allen/fig_additional_#3_#4_subject-5.pdf}
    \caption{Subject 5.}
  \end{subfigure}
  \hfill
  \begin{subfigure}[t]{0.49\linewidth}
    \centering
    \includegraphics[width=\linewidth]{figures/allen/fig_additional_#3_#4_subject-7.pdf}
    \caption{Subject 7.}
  \end{subfigure}
  \caption{\textbf{Per-subject #1 operating maps for #2 decoders on NSD.}
  Each subfigure reports raw and normalized accuracy, seed-averaged raw scaling curves, and relative improvement over matched real-only for one subject.}
  \label{fig:app-#3-#4-allen-subjects}
\end{figure}
}

\newcommand{\additionalboldsubjectmaps}[4]{%
\begin{figure}[H]
  \centering
  \begin{subfigure}[t]{0.49\linewidth}
    \centering
    \includegraphics[width=\linewidth]{figures/chang/fig_additional_#3_#4_subject-1.pdf}
    \caption{Subject 1.}
  \end{subfigure}
  \hfill
  \begin{subfigure}[t]{0.49\linewidth}
    \centering
    \includegraphics[width=\linewidth]{figures/chang/fig_additional_#3_#4_subject-2.pdf}
    \caption{Subject 2.}
  \end{subfigure}
  \begin{subfigure}[t]{0.49\linewidth}
    \centering
    \includegraphics[width=\linewidth]{figures/chang/fig_additional_#3_#4_subject-3.pdf}
    \caption{Subject 3.}
  \end{subfigure}
  \hfill
  \begin{subfigure}[t]{0.49\linewidth}
    \centering
    \includegraphics[width=\linewidth]{figures/chang/fig_additional_#3_#4_subject-4.pdf}
    \caption{Subject 4.}
  \end{subfigure}
  \caption{\textbf{Per-subject #1 operating maps for #2 decoders on BOLD5000.}
  Each subfigure reports raw and normalized accuracy, seed-averaged raw scaling curves, and relative improvement over matched real-only for one subject.}
  \label{fig:app-#3-#4-chang-subjects}
\end{figure}
}

\additionalnsdsubjectmaps{Top-1}{Ridge}{top1}{ridge}
\additionalboldsubjectmaps{Top-1}{Ridge}{top1}{ridge}
\additionalnsdsubjectmaps{Top-1}{Deep}{top1}{fmrimlp}
\additionalboldsubjectmaps{Top-1}{Deep}{top1}{fmrimlp}
\additionalnsdsubjectmaps{Top-5}{Ridge}{top5}{ridge}
\additionalboldsubjectmaps{Top-5}{Ridge}{top5}{ridge}
\additionalnsdsubjectmaps{Top-5}{Deep}{top5}{fmrimlp}
\additionalboldsubjectmaps{Top-5}{Deep}{top5}{fmrimlp}

\let\additionalaveragemap\relax
\let\additionalnsdsubjectmaps\relax
\let\additionalboldsubjectmaps\relax

\section{Additional Controls and Analyses}
\label{app:additional-analyses}

\paragraph{Shuffled control across the operating grid.}
Table~\ref{tab:shuffled-control} extends the Panel D Shuffled comparison to all retained real fractions and to $a\in\{1,2,4,8\}$.

\begin{table}[H]
  \centering
  \caption{\textbf{TRIBE versus Shuffled across the operating grid.}
  For each subject, the Top-10 accuracy difference between TRIBE and Shuffled is averaged over the retained real fractions $p\in\{10,30,50,70,90,100\}\%$.
  Differences are reported as mean $\pm$ SEM across four subjects per dataset.
  For every decoder and augmentation factor, the average difference is positive in all eight subjects.}
  \label{tab:shuffled-control}
  \begin{tabular}{cccc}
    \toprule
    Decoder & Aug. factor & NSD $\Delta$ (pp) & BOLD5000 $\Delta$ (pp) \\
    \midrule
    \multirow{4}{*}{Ridge} & $1\times$ & $+0.8 \pm 0.2$ & $+3.0 \pm 0.1$ \\
     & $2\times$ & $+1.9 \pm 0.2$ & $+5.4 \pm 0.1$ \\
     & $4\times$ & $+2.3 \pm 0.2$ & $+7.1 \pm 0.7$ \\
     & $8\times$ & $+2.6 \pm 0.2$ & $+7.0 \pm 1.1$ \\
    \midrule
    \multirow{4}{*}{Deep} & $1\times$ & $+3.1 \pm 0.2$ & $+7.4 \pm 0.5$ \\
     & $2\times$ & $+4.3 \pm 0.3$ & $+10.3 \pm 0.6$ \\
     & $4\times$ & $+3.5 \pm 0.4$ & $+10.9 \pm 1.0$ \\
     & $8\times$ & $+1.1 \pm 0.2$ & $+9.5 \pm 1.2$ \\
    \bottomrule
  \end{tabular}
\end{table}

\paragraph{Prediction-space bias--variance analysis.}
To probe why low-data regimes benefit while high augmentation factors hurt, we decomposed Ridge prediction error on held-out measured fMRI across five split seeds into seed variance and squared systematic prediction error.
In low-data regimes ($p=10$--$30\%$), very small augmentation factors slightly reduce seed variance, while moderate factors ($a=2$--$8$) primarily reduce systematic error, bringing the average prediction closer to the correct image embedding.
At the largest factor ($a=16$), both variance and systematic error increase relative to their best intermediate values.
In a high-data regime ($p=90\%$), augmentation does not reduce systematic error and total error can increase from small factors.
Together with the relatively weak synthetic-only baselines, these results are consistent with TRIBE providing useful stimulus--response structure when real data are scarce, while larger synthetic-data ratios increasingly expose the mismatch between TRIBE predictions and subject-specific responses.

\paragraph{Image-to-video sensitivity.}
TRIBE's visual pathway expects video, so our reference protocol repeats each image for 3\,s.
At the scarce-data Ridge operating point ($p=10\%$, $a=8$), we compare this protocol with a 5\,s repetition and a deterministic 3\,s pan/zoom.
Table~\ref{tab:videoization-control} shows that Top-10 retrieval remains similar across these choices.
We therefore do not claim that repeated-frame rendering is optimal; rather, the augmentation benefit is robust to these image-to-video conversions.

\begin{table}[H]
  \centering
  \caption{\textbf{Sensitivity to image-to-video conversion.}
  Top-10 is normalized to the corresponding 100\% real-only Ridge baseline.
  Values are mean $\pm$ SEM across four subjects at $p=10\%$ and $a=8$.}
  \label{tab:videoization-control}
  \begin{tabular}{llc}
    \toprule
    Dataset & Rendering & Top-10 (\% full-real) \\
    \midrule
    NSD & Repeat 3\,s & $43.4 \pm 1.6$ \\
    NSD & Pan/zoom 3\,s & $45.3 \pm 1.6$ \\
    NSD & Repeat 5\,s & $45.0 \pm 1.3$ \\
    BOLD5000 & Repeat 3\,s & $88.5 \pm 6.8$ \\
    BOLD5000 & Pan/zoom 3\,s & $86.8 \pm 3.4$ \\
    BOLD5000 & Repeat 5\,s & $89.0 \pm 3.7$ \\
    \bottomrule
  \end{tabular}
\end{table}

\section{Image reconstruction with DynaDiff}
\label{app:dynadiff}

To adapt DynaDiff to be trained on both synthetic and real fMRI data, we slightly modify the brain module architecture in the following way: i) We add an initial linear layer to process independently real or synthetic data (Modality-specific initial layer), and ii) As synthetic fMRI data contains only one TR, we only use the last temporal aggregation layer for real data and keep only the first TR layer for synthetic fMRI data (Modality-specific temporal aggregation). All other architectural choices and training hyperparameters follow the original DynaDiff recipe~\citep{careil2025dynadiff}, including LoRA fine-tuning of the diffusion U-Net. Each run was trained on 8 NVIDIA H100 GPUs for approximately 2 days.

\begin{figure}[H]
  \centering
  \includegraphics[width=\linewidth]{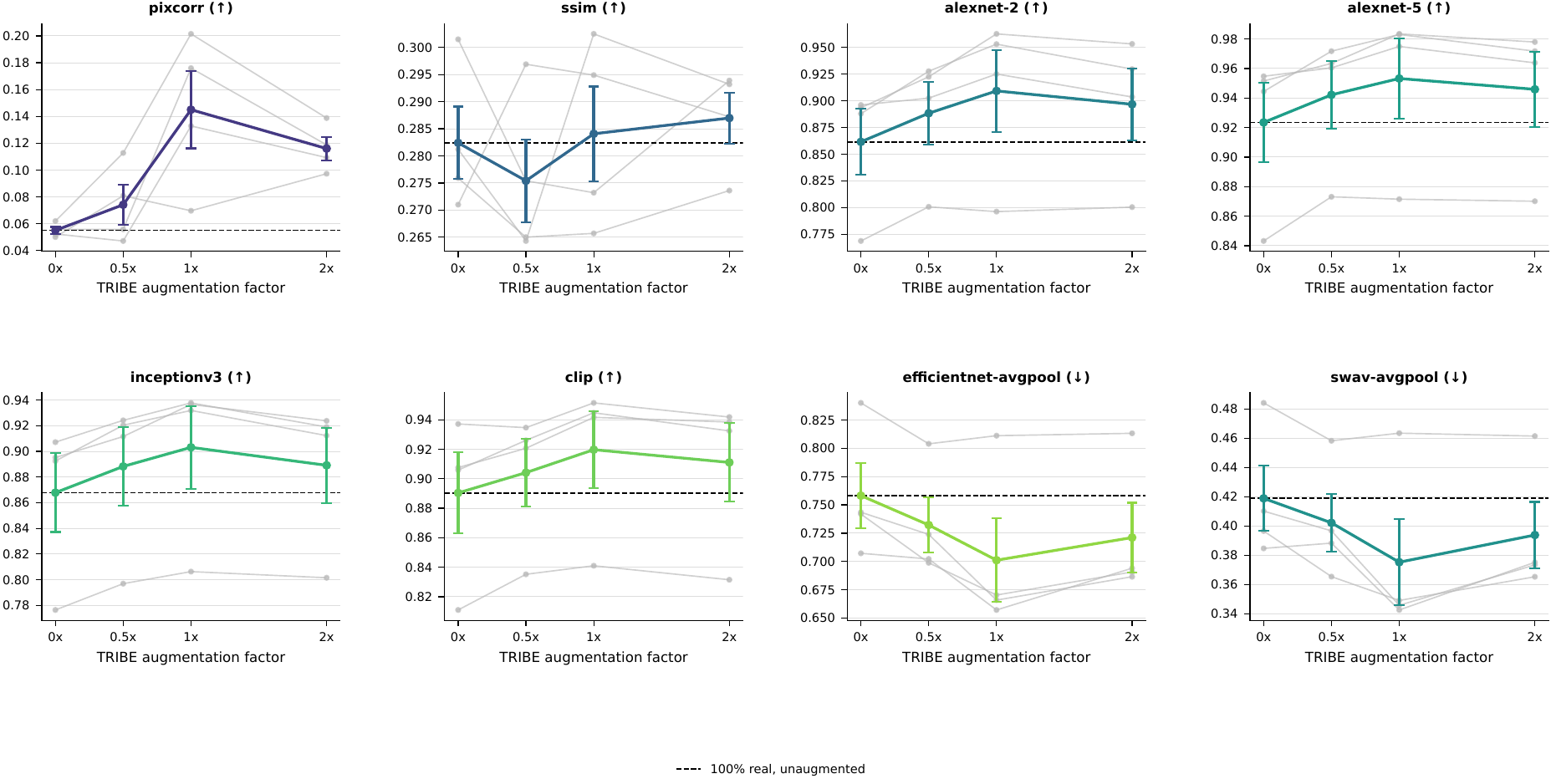}
  \caption{\textbf{All reconstruction metrics across four NSD subjects.}
  Colored curves and error bars report mean $\pm$ SEM; gray curves show individual subjects.
  The dashed line is the mean 100\% real, unaugmented reference.
  Seven of eight metrics improve for all four subjects at $a=1$, while SSIM changes little on average.}
  \label{fig:dynadiff-all-metrics}
\end{figure}

Figure~\ref{fig:dynadiff-all-metrics} reports all eight reconstruction metrics and their per-subject trajectories.
At $a=1$, all eight subject-averaged metrics improve over the real-only baseline, and every metric except SSIM improves for all four subjects.
Subject 7 shows lower absolute reconstruction performance than the other subjects across most metrics, consistent with the single-subject results reported by MindEye2~\citep{scotti2024mindeye2}; this suggests that the difference is not specific to our augmentation analysis.

\section{Scan-Time Interpretation of Operating Grids}
\label{app:scan-time}

Tables~\ref{tab:nsd-scan-time} and~\ref{tab:bold5000-scan-time} translate the abstract operating grid coordinates into approximate scan-time budgets.
For a dataset with real-fMRI training budget $H$, retaining $p\%$ real data corresponds to $(p/100)H$ real scan hours, while augmentation factor $a$ adds $a(p/100)H$ synthetic recording-equivalent hours.
Only the first quantity is acquired in the scanner; the second is generated by TRIBE.
This framing separates acquisition cost from effective training-set scale.

\begin{table}[H]
  \centering
  \scriptsize
  \caption{\textbf{NSD scan-time interpretation of the operating grid.} Each cell reports real scan hours + synthetic recording-equivalent hours for the retained-real percentage $p$ and augmentation factor $a$. Synthetic hours are not acquired; they express how much additional stimulus-conditioned fMRI the decoder sees if one TRIBE prediction is treated as one recording-equivalent image response.}
  \label{tab:nsd-scan-time}
  \resizebox{\linewidth}{!}{%
  \begin{tabular}{lccccccccc}
    \toprule
    $p$ & $a=0$ & $a=0.25$ & $a=0.5$ & $a=0.75$ & $a=1$ & $a=2$ & $a=4$ & $a=8$ & $a=16$ \\
    \midrule
    10\% & 1+0 & 1+0.2 & 1+0.5 & 1+0.8 & 1+1 & 1+2 & 1+4 & 1+8 & 1+16 \\
    30\% & 3+0 & 3+0.8 & 3+1.5 & 3+2.2 & 3+3 & 3+6 & 3+12 & 3+24 & 3+48 \\
    50\% & 5+0 & 5+1.2 & 5+2.5 & 5+3.8 & 5+5 & 5+10 & 5+20 & 5+40 & 5+80 \\
    70\% & 7+0 & 7+1.8 & 7+3.5 & 7+5.2 & 7+7 & 7+14 & 7+28 & 7+56 & 7+112 \\
    90\% & 9+0 & 9+2.2 & 9+4.5 & 9+6.8 & 9+9 & 9+18 & 9+36 & 9+72 & 9+144 \\
    100\% & 10+0 & 10+2.5 & 10+5 & 10+7.5 & 10+10 & 10+20 & 10+40 & 10+80 & 10+160 \\
    \bottomrule
  \end{tabular}%
  }
\end{table}

\begin{table}[H]
  \centering
  \scriptsize
  \caption{\textbf{BOLD5000 scan-time interpretation of the operating grid.} Each cell reports real scan hours + synthetic recording-equivalent hours for the retained-real percentage $p$ and augmentation factor $a$. Synthetic hours are not acquired; they express how much additional stimulus-conditioned fMRI the decoder sees if one TRIBE prediction is treated as one recording-equivalent image response.}
  \label{tab:bold5000-scan-time}
  \resizebox{\linewidth}{!}{%
  \begin{tabular}{lccccccccc}
    \toprule
    $p$ & $a=0$ & $a=0.25$ & $a=0.5$ & $a=0.75$ & $a=1$ & $a=2$ & $a=4$ & $a=8$ & $a=16$ \\
    \midrule
    10\% & 1.3+0 & 1.3+0.3 & 1.3+0.7 & 1.3+1.0 & 1.3+1.3 & 1.3+2.7 & 1.3+5.3 & 1.3+10.7 & 1.3+21.3 \\
    30\% & 4.0+0 & 4.0+1.0 & 4.0+2.0 & 4.0+3.0 & 4.0+4.0 & 4.0+8.0 & 4.0+16.0 & 4.0+32.0 & 4.0+64.0 \\
    50\% & 6.7+0 & 6.7+1.7 & 6.7+3.3 & 6.7+5.0 & 6.7+6.7 & 6.7+13.3 & 6.7+26.7 & 6.7+53.4 & 6.7+106.7 \\
    70\% & 9.3+0 & 9.3+2.3 & 9.3+4.7 & 9.3+7.0 & 9.3+9.3 & 9.3+18.7 & 9.3+37.4 & 9.3+74.7 & 9.3+149.4 \\
    90\% & 12.0+0 & 12.0+3.0 & 12.0+6.0 & 12.0+9.0 & 12.0+12.0 & 12.0+24.0 & 12.0+48.0 & 12.0+96.0 & 12.0+192.1 \\
    100\% & 13.3+0 & 13.3+3.3 & 13.3+6.7 & 13.3+10.0 & 13.3+13.3 & 13.3+26.7 & 13.3+53.4 & 13.3+106.7 & 13.3+213.4 \\
    \bottomrule
  \end{tabular}%
  }
\end{table}

\section{Dataset Licenses}
\label{app:dataset-licenses}

We used the Natural Scenes Dataset (NSD)~\citep{allen2022massive} and BOLD5000~\citep{chang2019bold5000} in accordance with their respective license terms and conditions.
For both datasets, we accessed the data through the official distribution channels, agreed to the applicable terms of use, and used the data only for research purposes consistent with those terms.

\end{document}